\documentclass{article} 
\usepackage{iclr2021_conference,times}

\usepackage{amsmath,amsfonts,bm}

\newcommand{\red}[1]{\textcolor{red}{#1}}

\definecolor{green}{rgb}{0.0, 0.6, 0.3}
\newcommand{\green}[1]{\textcolor{green}{#1}}

\definecolor{mypink3}{cmyk}{0, 0.7808, 0.4429, 0.1412}

\definecolor{mypink2}{RGB}{0, 100, 250}

\newcommand{\ie}{{\em i.e.,}\ }
\newcommand{\eg}{{\em e.g.,}\ }

\newcommand{\wrt}{\emph{w.r.t.}\ }

\newcommand{\Ni}{({\em i})~}
\newcommand{\Nii}{({\em ii})~}
\newcommand{\Niii}{({\em iii})~}
\newcommand{\Niv}{({\em iv})~}

\def\eqref#1{equation~\ref{#1}}

\def\1{\bm{1}}

\def\valpha{{\bm{\alpha}}}
\def\vtau{{\bm{\tau}}}

\def\vs{{\bm{s}}}

\def\vw{{\bm{w}}}

\def\mA{{\bm{A}}}

\def\mP{{\bm{P}}}
\def\mQ{{\bm{Q}}}

\def\mS{{\bm{S}}}

\def\mW{{\bm{W}}}

\DeclareMathAlphabet{\mathsfit}{\encodingdefault}{\sfdefault}{m}{sl}
\SetMathAlphabet{\mathsfit}{bold}{\encodingdefault}{\sfdefault}{bx}{n}
\newcommand{\tens}[1]{\bm{\mathsfit{#1}}}
\def\tA{{\tens{A}}}

\def\tS{{\tens{S}}}

\def\gL{{\mathcal{L}}}

\def\gT{{\mathcal{T}}}

\newcommand{\E}{\mathbb{E}}

\newcommand{\softmax}{\mathrm{softmax}}

\usepackage{hyperref}

\usepackage{url}
\usepackage{times}
\usepackage{graphicx}
\usepackage{breqn}
\usepackage{bbm}
\usepackage{amsmath,bm}
\usepackage{latexsym}
\usepackage{multirow}
\usepackage{enumitem}
\usepackage{amssymb}
\usepackage{pifont}
\usepackage[colorinlistoftodos,prependcaption,textsize=tiny]{todonotes}

\newcommand*{\ditto}{---\texttt{"}---}
\newcommand{\cmark}{\ding{51}}%
\newcommand{\xmark}{\ding{55}}%

\newlength\myindent
\newcommand\bindent{%
  \begingroup
  \setlength{\itemindent}{\myindent}
  \addtolength{\algorithmicindent}{\myindent}
}
\newcommand\eindent{\endgroup}
\usepackage{caption}
\usepackage{algorithm}
\usepackage{authblk}
\usepackage{algorithmic}
\usepackage{subcaption}
\usepackage{wrapfig}
\usepackage{booktabs}
\usepackage{comment}
\def\shortname{WNSMN}
\def\longname{Weakly Supervised Neuro-Symbolic Module Network~}
\def\longnamebold{\textbf{W}eakly-Supervised
\textbf{N}euro-\textbf{S}ymbolic  \textbf{M}odule \textbf{N}etwork}

\usepackage{cleveref}
\crefformat{section}{\S#2#1#3} 
\crefformat{subsection}{\S#2#1#3}
\crefformat{subsubsection}{\S#2#1#3}

\usepackage{microtype}

\title{Weakly Supervised Neuro-Symbolic Module Networks for Numerical Reasoning}

\author[*]{Amrita Saha}
\author[*+]{Shafiq Joty}
\author[*]{Steven C.H. Hoi}
\affil[*]{Salesforce AI Research}
\affil[+]{Salesforce AI Research}
\affil[ ]{\textit {\{amrita.saha, sjoty, shoi\}@salesforce.com}}
\iclrfinalcopy 
\begin{document}

\maketitle

\begin{abstract}
Neural Module Networks (NMNs) have been quite successful in incorporating explicit reasoning as learnable modules in various question answering tasks, including the most generic form of numerical reasoning over text  in Machine Reading Comprehension (MRC). However, to achieve this, contemporary NMNs need strong supervision in executing the query as a specialized program over reasoning modules and fail to generalize to more open-ended settings without such supervision. Hence we propose {\longnamebold} (\shortname) trained with answers as the sole supervision for numerical reasoning based MRC. It learns to execute a noisy heuristic program obtained from the dependency parsing of the query, as discrete actions over both neural and symbolic reasoning modules and trains it end-to-end in a reinforcement learning framework with discrete reward from answer matching. On the numerical-answer subset of DROP, {\shortname} outperforms NMN by 32\% and the reasoning-free language model GenBERT by 8\% in exact match accuracy when trained under comparable weak supervised settings. This showcases the effectiveness and generalizability of modular networks that can handle explicit discrete reasoning over noisy programs in an end-to-end manner. 
\end{abstract}

\section{Introduction} \label{sec:intro}

\begin{figure*}[t!]
\centering
\includegraphics[width=\textwidth]{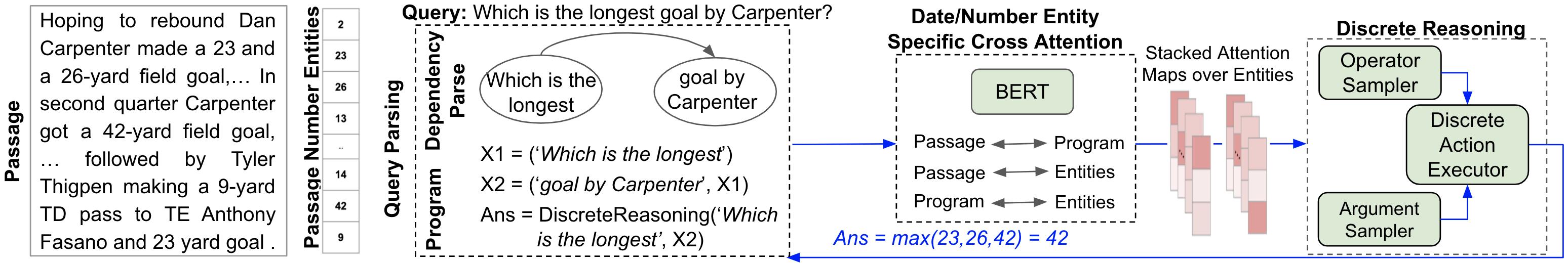}
\caption{Example (passage, query, answer) from DROP and outline of our method: executing noisy program obtained from dependency parsing of query by learning date/number entity specific cross attention, and sampling and execution of discrete operations on entity arguments to reach the answer.}
\label{fig:example}
\end{figure*}

End-to-end neural models have proven to be powerful tools for an expansive set of language and vision problems by effectively emulating the \emph{input-output} behavior. However, many real problems like Question Answering (QA) or Dialog need more interpretable models that can incorporate explicit reasoning in the inference. In this work, we focus on the most generic form of numerical reasoning over text, encompassed by the reasoning-based MRC framework. A particularly challenging setting for this task is where the answers are numerical in nature as in the popular MRC dataset, DROP \citep{Dua2019DROP}. 
Figure \ref{fig:example} shows the intricacies involved in the task, \Ni passage and query language understanding, \Nii contextual understanding of the passage date and numbers, and \Niii application of quantitative reasoning (\eg\ \emph{max, not}) over dates and numbers to reach the final numerical answer.

Three broad genres of models have proven successful on the DROP numerical reasoning task. 
\\
First, \emph{large-scale pretrained language models} like GenBERT \citep{Geva2020InjectingNR} uses a monolithic Transformer architecture and decodes numerical answers digit-by-digit. {Though they deliver mediocre performance when trained only on the target data, their competency is derived from pretraining on massive synthetic data augmented with explicit supervision of the gold numerical reasoning.} 
\\
Second kind of models are the \emph{reasoning-free hybrid models} like NumNet \citep{ran-etal-2019-numnet}, NAQANet \citep{Dua2019DROP}, NABERT+ \citep{nabert} and MTMSN \citep{hu2019multi}, NeRd \citep{DBLP:conf/iclr/ChenLYZSL20}. They explicitly incorporate numerical computations in the standard extractive QA pipeline by learning a multi-type answer predictor over different reasoning types (\eg \emph{max/min}, \emph{diff/sum}, \emph{count}, \emph{negate}) and directly predicting the corresponding numerical expression, instead of learning to reason. This is facilitated by exhaustively precomputing all possible outcomes of discrete operations and augmenting the training data with the reasoning-type supervision and 
numerical expressions that lead to the correct answer. 
\\
Lastly, the most relevant class of models to consider for this work are the \emph{modular networks for reasoning}. Neural Module Networks (NMN) \citep{Gupta2020Neural} is the first explicit reasoning based QA model which parses the query into a specialized program and executes it step-wise over learnable reasoning modules. However, to do so, apart from the exhaustive precomputation of all discrete operations, it also needs more fine-grained supervision of the gold program and the gold program execution, obtained heuristically, by leveraging the abundance of templatized queries in DROP.

While being more pragmatic and richer at interpretability, both modular and hybrid networks are also tightly coupled with the additional supervision. For instance, the hybrid models cannot learn without it, and while NMN is the first to \emph{enable} learning from QA pair alone, it still needs more finer-grained supervision for at least a part of the training data. With this, it manages to supercede the SoTA models NABERT and MTMSN on a carefully chosen subset of DROP using the supervision. However, NMN generalizes poorly to more open-ended settings where such supervision is not easy to handcraft. 


\textbf{Need for symbolic reasoning.} One striking characteristic of the modular methods is to avoid discrete reasoning by employing only learnable modules with an exhaustively precomputed space of outputs. While they perform well on DROP, their modeling complexity grows arbitrarily with more complex non-linear numerical operations (\eg\ {$\exp$, $\log$, $\cos$}). 
Contrarily, symbolic modular networks that execute the discrete operations are possibly more robust or pragmatic in this respect by remaining unaffected by the operation complexity.
Such discrete reasoning has indeed been incorporated for simpler, well-structured tasks like math word problems \citep{koncel-kedziorski-etal-2016-mawps} or KB/Table-QA \citep{zhong2017seq2sql,NIPS2018_8204,saha-etal-2019-complex}, with Deep Reinforcement Learning (RL) for end-to-end training. MRC however needs a more generalized framework of modular neural networks involving more fuzzy reasoning over noisy entities extracted from open-ended passages. 

In view of this, we propose a \longnamebold~\textbf{(\shortname)}
\vspace{-1em}
\begin{itemize}[leftmargin=*]
    \item A first attempt at numerical reasoning based MRC, trained with answers as the sole supervision;
    \vspace{-0.4em}
    \item Based on a generalized framework of dependency parsing of queries into noisy heuristic programs;
    \vspace{-0.4em}\item  End-to-end training of neuro-symbolic reasoning modules in a  RL framework with discrete rewards;
\end{itemize}

To concretely compare \shortname\ with contemporary NMN, consider the example in \Cref{fig:example}. {In comparison to our generalized query-parsing,} NMN parses the query into a program form {\small\emph{(MAX(FILTER(FIND(`Carpenter'), `goal'))}}, which is step-wise executed by different learnable modules with exhaustively precomputed output set. To train the network, it employs various forms of strong supervision such as gold program operations and gold query-span attention at each step of the program and gold execution \ie\ supervision of the passage numbers ({\small\emph{23, 26, 42}}) to execute {\small\emph{MAX}} operation on.

While NMN can only handle the 6 reasoning categories that the supervision was tailored to, {\shortname} focuses on the full DROP with numerical answers (called DROP-\emph{num}) that involves more diverse reasoning on more open-ended questions.
We empirically compare {\shortname} on DROP-\emph{num} with the SoTA NMN and GenBERT that allow learning with partial or no strong supervision. Our results showcase that the proposed {\shortname} achieves 32\% better accuracy than NMN in absence of at least one or more types of supervision and performs 8\% better than GenBERT when the latter is fine-tuned only on DROP in a comparable setup, without additional synthetic data having explicit supervision. 


\section{Model: \longname}
\label{sec:model}


We now describe our proposed  {\shortname} that learns to infer the answer based on weak supervision of the QA pair by generating the program form of the query and executing it through explicit reasoning.

\paragraph{Parsing Query into Programs} To keep the framework generic, we use a simplified representation of the Stanford dependency parse tree \citep{chen-2014-fast} of the query to get a generalized program (\Cref{appendix:query_parse}). First, a node is constructed for the subtree rooted at each child of the root by merging its descendants in the original word order. Next an edge is added from the left-most node (which we call the \emph{root clause}) to every other node. Then by traversing left to right, each node is organized into a step of a program having a linear flow. For example, the program obtained in Figure \ref{fig:example} is 
{\small 
\emph{X1 = (`which is the longest')}; ~\emph{X2 = (`goal by Carpenter', X1)}; ~\emph{Answer = Discrete-Reasoning(`which is the longest', X2)}}. Each program step consists of two types of arguments \Ni Query Span Argument obtained from the corresponding node, indicates the query segment referred to, in that program step \eg\ {\small\emph{`goal by Carpenter'}} in Step 2 (ii) Reference Argument(s) obtained from the incoming edges to that node, refers to the previous steps of the program that the current one depends on  \eg {\small\emph{X1}} in Step 2.
Next, a final step of the program is added, which has the reference argument as the leaf node(s) obtained in the above manner and the query span argument as the root-clause. This step is specifically responsible for handling the discrete operation,  enabled by the root-clause which is often indicative of the kind of discrete reasoning involved (\eg\ {\small\emph{max}}). However this being a noisy heuristic, the QA model needs to be robust to such noise and additionally rely on the full query representation in order to predict the discrete operation. For simplicity we limit the number of reference arguments to 2.

\subsection{Program Execution}

Our proposed {\shortname} learns to execute the program over the passage in three steps. In the preprocessing step, it identifies numbers and dates from the passage, and maintains them as separate canonicalized entity-lists along with their mention locations. Next, it learns an entity-specific \emph{cross-attention} model to rank the entities \wrt their query-relevance (\Cref{subsec:extraction}), and then \emph{samples} relevant entities as discrete arguments (\Cref{subsec:select}) and executes appropriate discrete operations on them to reach the answer. An RL framework (\Cref{subsec:RL}) trains it end-to-end with the answer as the sole supervision.





\subsubsection{Entity-Specific Cross Attention for Information Extraction} \label{subsec:extraction}

To rank the query-relevant passage entities, we model the  passage, program and entities jointly.

\begin{figure}[t!]
\centering
\includegraphics[width=\textwidth]{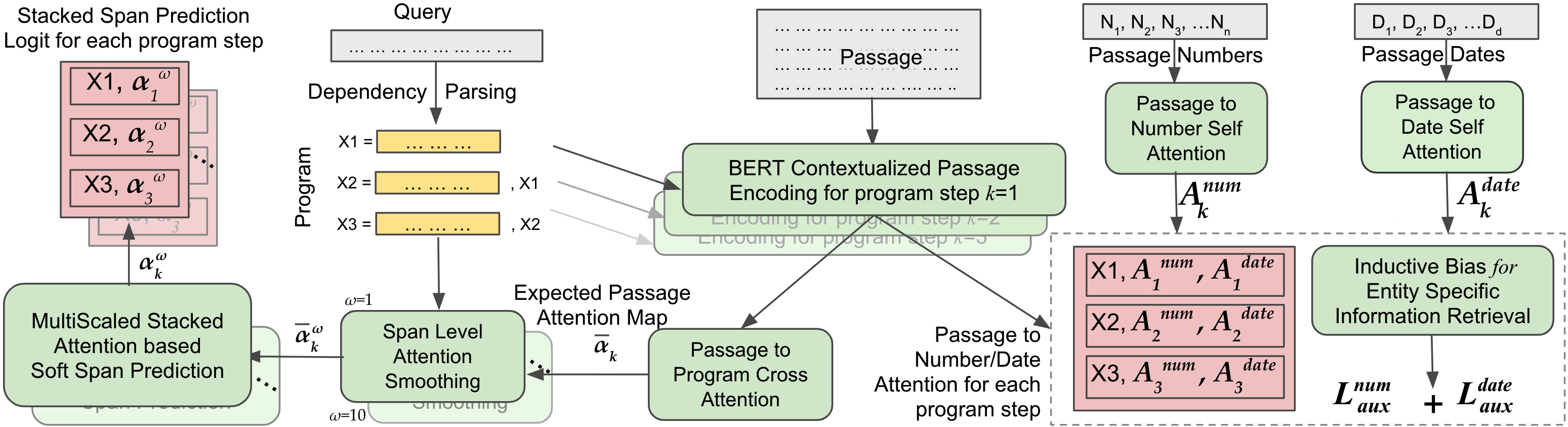}
\caption{Modeling the interaction between the passage and (left) the program, \& (right) its number/date entities. For each program step $k$, they respectively yield (i) Stacked Span Prediction Logits and (ii) Attention over Number/Date entities for each passage token. The linear combination of these two gives the expected distribution over entities, $\mathcal{T}^{num}_k$ and $\mathcal{T}^{date}_k$ for the step $k$} 
\label{fig:passage_entities}
\end{figure}

\paragraph{Modeling interaction between program and passage} This module (\Cref{fig:passage_entities}, left) learns to associate \emph{query span} arguments of the program with the passage. For this, similar to NMN, we use a BERT-base pretrained encoder \citep{devlin2018pretraining} to get contextualized token embeddings of the passage and query span argument of each program step, respectively denoted by {$\mP_k$} and $\mQ_k$ for the $k$'th program step. Based on it, we learn a \emph{similarity} matrix $\tS \in \mathbbm{R}^{l\times n\times m}$ between the program and passage, where $l$, $n$, and $m$ respectively are the program length and query span argument and passage length (in tokens). Each $\mS_k \in \mathbbm{R}^{n\times m}$ represents the affinity over the passage tokens for the $k$'th program argument and is defined as $\mS_k(i,j) = \vw^T [\mQ_{ki} ; \mP_{kj} ; \mQ_{ki} \odot \mP_{kj}]$, where $\vw$ is a learnable parameter and $\odot$ is element-wise multiplication. From this, an attention map $\mA_k$ is computed over the passage tokens for the $k$'th program argument as $\mA_k (i,j)  = \softmax_j(\mS_{k}(i,j)) = \frac{\exp (\mS_k (i,j)) }{ \sum_j \exp (\mS_k(i,j)) }$. 
\noindent Similarly, for the $i$'th token of the $k$'th program argument the cumulative attention $a_{ki}$ \wrt the passage is $a_{ki}=\softmax_i(\sum_j \mS_{k}(i,j))$. A linear combination of the attention map $\mA_k(i, \cdot)$ weighted by $a_{ki}$ 
gives the expected passage attention for the $k$'th step, $\bar{\valpha}_k = \sum_i a_{ki} \mA_k(i, \cdot) \in \mathbbm{R}^{m}$.

\textit{Span-level smoothed attention.} To facilitate information spotting and extraction over contiguous spans of text, we regularize the passage attention so that the attention on a passage token is high if the attention over its neighbors is so. We achieve this by adopting a heuristic smoothing technique \citep{DBLP:journals/ijon/HuangDSC20}, taking a sliding window of different lengths $\omega = \{1,2, \ldots 10\}$ over the passage, and replacing the token-level attention with the attention averaged over the window. This results in $10$ different attention maps over the passage for the $k$'th step of the program: 
$\{\bar{\valpha}^{\omega}_k | \omega \in \{1,2, $\ldots$, 10\}\}$.


\textit{Soft span prediction.} This network takes a multi-scaled \citep{Gupta2020Neural} version of $\bar{\valpha}^{\omega}_k$ , by multiplying the attention map with $|\vs|$ different scaling factors ($\vs = \{1,2,5,10\}$), yielding a $|\vs|$-dimensional representation for each passage token, \ie\ $\bar{\valpha}^{\omega}_{k} \in \mathbbm{R}^{m \times |\vs|}$.  This is then passed through a $L$-layered stacked self-attention \emph{transformer} block \citep{NIPS2017_7181}, which encodes it to $m\times d$ dimension, followed by a \emph{linear layer} of dimension $d\times 1$, to obtain the span prediction logits: $\valpha^{\omega}_k = Linear(Transformer(MultiScaling(\bar{\valpha}^{\omega}_k))  \in \mathbbm{R}^{m} $. Further the span prediction logits at each program step (say $k$) is additively combined with those from the previous steps referenced in the current one, through the reference argument ($ref(k)$) at step $k$, \ie $\valpha^{\omega}_k =  \valpha^{\omega}_k + \sum_{k' \in ref(k)} \valpha^{\omega}_{k'}$.
\paragraph{Modeling interaction between program and number/date entities}

This module (\Cref{fig:passage_entities}, right) facilitates an entity-based information spotting capability, that is, given a passage mention of a number/date entity relevant to the query, the model should be able to attend to the neighborhood around it.
To do this, for each program step, we first compute a \emph{passage tokens to number tokens} attention map $\tA^{num} \in \mathbbm{R}^{l\times m\times N}$, where  $N$ is the number of unique number entities. Note that this attention map is different for each program step as the contextual BERT encoding of the passage tokens  ($\mP_k$) is coupled with the program's span argument of that step. At the $k$-th step, the row $\mA^{num}_{k}(i,\cdot)$ denotes the probability distribution over the $N$ unique number tokens \wrt\ the $i$-th passage token. The attention maps are obtained by a softmax normalization of each row of the corresponding \emph{passage tokens to number tokens} similarity matrix, $\mS^{num}_k \in \mathbbm{R}^{m\times N}$ for $k =\{ 1 \ldots l\}$, where the elements of $\mS^{num}_k$ are computed as $\mS^{num}_{k}(i,j) = \mP_{ki}^T \mW_{n}\mP_{kn_j}$ with $\mW_{n} \in \mathbbm{R}^{d\times d}$ being a learnable projection matrix and $n_j$ being the passage location of the $j$-th number token. These similarity scores are additively aggregated over all mentions of the same number entity in the passage. 

The relation between program and entities is then modeled as $\vtau^{\omega}_k = \softmax(\sum_i {\alpha}^{\omega}_{ki} \mA^{num}_k(i,\cdot))\in \mathbbm{R}^{N}$, which gives the expected distribution over the $N$ number tokens for the $k$-th program step and using $\omega$ as the smoothing window size. The final stacked attention map obtained for the different windows is $\gT^{num}_k = \{ \vtau^{\omega}_k | \omega \in \{1,2,\ldots 10\}\}$. Similarly, for each program step $k$, we also compute a separate stacked attention map $\gT^{date}_k$ over the unique date tokens, parameterized by a different $\mW_{d}$.

A critical requirement for a meaningful attention over entities is to incorporate information extraction capability in the number and date attention maps $\tA^{num}$ and $\tA^{date}$, by enabling the model to attend over the neighborhood of the relevant entity mentions. This is achieved by minimizing the unsupervised auxiliary losses $\gL^{num}_{aux}$ and $\gL^{date}_{aux}$ in the training objective, which impose an inductive bias over the number and date entities, similar to \cite{Gupta2020Neural}. Its purpose is to ensure that the passage attention is densely distributed inside the neighborhood of $\pm~\Omega$ (a hyperparameter, \eg 10) of the passage location of the entity mention, without imposing any bias on the attention distribution outside the neighborhood. 
Consequently, it maximises the log-form of cumulative likelihood of the attention distribution inside the window and the entropy of the attention distribution outside of it.
\begin{dmath}
\gL^{num}_{aux} = -\frac{1}{l}\sum\limits_{k=1}^{l}\bigg{[}\sum\limits_{i=1}^{m} \big{[}\log(\sum\limits_{j=1}^{N}\displaystyle{\mathbbm{1}}_{n_j\in[i\pm~\Omega]}a^{num}_{kij}) - \sum\limits_{j=1}^{N} \displaystyle{\mathbbm{1}}_{n_j\not\in[i\pm~\Omega]}a^{num}_{kij}\log (a^{num}_{kij})\big{]}\bigg{]} \label{eq:unsup-loss}
\end{dmath}

where $\displaystyle{\mathbbm{1}}$ is indicator function and $a^{num}_{kij}=\mA^{num}_{k}(i,j)$. $\gL^{date}_{aux}$ for date entities is similarly defined.

\subsubsection{Modeling Discrete Reasoning} \label{subsec:select}

\begin{figure}[t!]
\centering
\includegraphics[width=\textwidth]{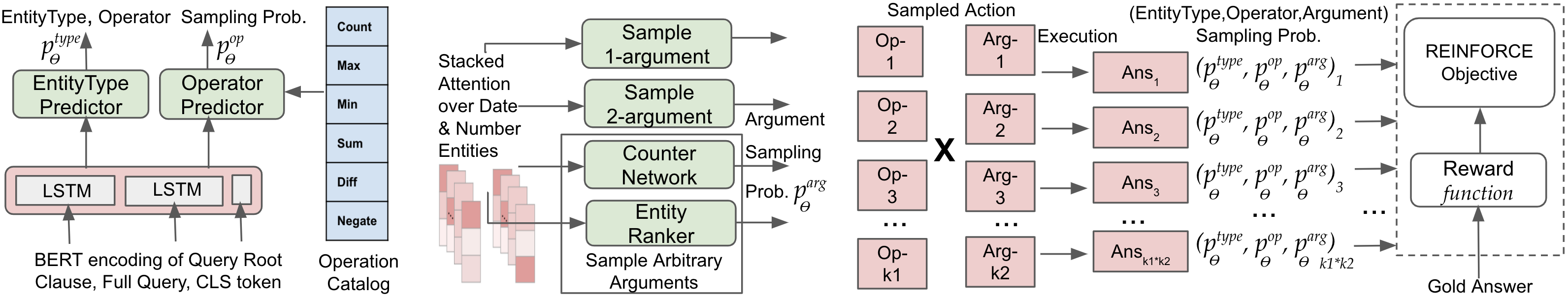}
\caption{Operator \& Argument Sampling Network and RL framework over sampled discrete actions}
\label{fig:operator}
\end{figure}

The model next learns to execute a single step\footnote{This is a reasonable assumption  for DROP with a recall of 90\% on the training set. However, it does not limit the generalizability of \shortname, as with standard beam search it is possible to scale to an $l$-step MDP.} of discrete reasoning (\Cref{fig:operator}) based on the final program step. The final step contains \Ni root-clause of the query which often indicates the type of discrete operation (\eg\  \emph{`what is the longest'} indicates \texttt{max}, \emph{`how many goals'} indicates \texttt{count}), and \Nii reference argument indicating the previous program steps the final step depends on. Each previous step (say $k$) is represented as stacked attention maps $\mathcal{T}^{num}_k$ and    $\mathcal{T}^{date}_k$, obtained from \Cref{subsec:extraction}.  

\paragraph{Operator Sampling Network} 
Owing to the noisy nature of the program, the operator network takes as input: \Ni BERT's \texttt{[CLS]} representation for the passage-query pair and LSTM \citep{hochreiter1997long}  encoding (randomly initialized) of the BERT contextual representation of \Nii the root-clause from the final program step and \Niii full query (\wrt passage), to make two predictions:
\begin{itemize}[leftmargin=*]
\item \textit{Entity-Type Predictor Network}, an Exponential Linear Unit (Elu) activated {fully-connected layer} followed by a $\softmax$ that outputs the probabilities of sampling either date or number types. 
\item \textit{Operator Predictor Network}, a similar Elu-activated fully connected layer followed by a $\softmax$ which learns a probability distribution over a fixed catalog of 6 numerical and logical operations (\texttt{count}, \texttt{max}, \texttt{min}, \texttt{sum}, \texttt{diff}, \texttt{negate}), each represented with learnable embeddings. 
\end{itemize}
Apart from the \texttt{diff} operator which acts only on two arguments, all other operations can take any arbitrary number of arguments. Also some of these operations can be applied only on numbers (\eg\ \texttt{sum}, \texttt{negate}) while others can be applied on both numbers or date (\eg\ \texttt{max}, \texttt{count}).
\paragraph{Argument Sampling Network} 
This network learns to sample date/number entities as arguments for the sampled discrete operation, given the entity-specific stacked attentions ($\mathcal{T}^{num}_k$ and  $\mathcal{T}^{date}_k$) for each previous step (say, $k$), that appears in the reference argument of the final program step. 
In order to allow sampling of fixed or arbitrary number of arguments, the argument sampler learns four types of networks, each modeled with a $L$-layered stacked self attention based $Transformer$ block (with output dimension $d$) followed by different non-linear layers embodying their functionality and a $\softmax$ normalization to get the corresponding probability of the argument sampling (\Cref{fig:operator}). 
\begin{itemize}[leftmargin=*]
\item Sample $n \in \{1,2\}$ Argument Module: $\softmax(Elu(Linear_{d\times n}( Transformer(\mathcal{T}))))$, outputs a distribution over the single entities ($n=1$) or a joint distribution over the entity-pairs ($n=2$).
\item Counter Module: $\softmax(Elu(Linear_{d\times10}(CNN\text{-}Encoder(Transformer(\mathcal{T})))))$, predicts a distribution over possible count values ($\in[1,\ldots,10]$) of number of entity arguments to sample.
\item Entity-Ranker Module: $ \softmax(PRelu(Linear_{d\times1}(Transformer(\mathcal{T}))))$, learns to rerank the entities and outputs a distribution over all the entities given the stacked attention maps as input.
\item Sample Arbitrary Argument: $Multinomial$(Entity-Ranked  Distribution, Counter Prediction).
\end{itemize}

Depending on the number of arguments needed by the discrete operation and the number of reference arguments in the final program step, the model invokes one of  \emph{Sample \{1, 2, Arbitrary\} Argument}. For instance, if the sampled operator is \texttt{diff} which needs 2 arguments, and the final step has 1 or 2 reference arguments, then the model respectively invokes either \emph{Sample 2 argument} or \emph{Sample 1 argument} on the stacked attention $\mathcal{T}$ corresponding to each reference argument. And, for operations needing arbitrary number of arguments, the model invokes the \emph{Sampling Arbitrary Argument}. For the \emph{Arbitrary Argument} case,
the model first predicts the number of entities $\mathit{c} \in \{1, \ldots, 10\}$ to sample using the Counter Network, and then samples from the multinomial distribution based on the joint of $\mathit{c}$-combinations of entities constructed from the output distribution of the Entity Ranker module.


\subsubsection{Training with Weak Supervision in the Deep RL Framework}
\label{subsec:RL}
We use an RL framework to train the model with only discrete binary feedback from the exact match of the gold and predicted numerical answer. In particular, we use the REINFORCE \citep{Williams:92} policy gradient method where a stochastic policy comprising a sequence of actions is learned with the goal of maximizing the expected reward. In our case, the discrete operation along with argument sampling constitute the \emph{action}. However, because of our assumption that a single step of discrete reasoning suffices for most questions in DROP, we further simplify the RL framework to a contextual multi-arm bandit (MAB) problem with a 1-step MDP, \ie\ the agent performs only one step action. 
 
\vspace{-0.2em}
Despite the simplifying assumption of 1-step MDP, the following characteristics of the problem render it highly challenging: \Ni the action space $\mathcal{A}$ is exponential in the order of number of operations and argument entities in the passage (averaging to \emph{12K} actions for DROP-\emph{num}); 
\Nii the extreme reward sparsity owing to the binary feedback is further exacerbated by the presence of spurious rewards, as the same answer can be generated by multiple diverse actions. Note that previous approaches like NMN can avoid such spurious supervision because they heuristically obtain additional annotation of the question category, the gold program or gold program execution atleast for some training instances.  




In our contextual MAB framework, for an input $x$ = (passage($p$), query($q$)), the context or environment state $s_{\phi}(x)$ is modeled by the entity specific cross attention (\Cref{subsec:extraction}, parameterized by $\phi$) between the \Ni passage \Nii program-form of the query and \Niii extracted passage date/number entities. Given the state $s_{\phi}(x)$, the layout policy (\cref{subsec:select}, parameterized by $\theta$) then learns the query-specific inference layout, \ie\ the discrete action sampling policy $P_\theta(a|s_{\phi}(x))$ for action $a \in \mathcal{A}$. The action sampling probability is a product of the probability of sampling entities from the appropriate entity type ($P_{\theta}^{type}$), probability of sampling the operator ($P_{\theta}^{op}$), and probability of sampling the entity argument(s) ($P_{\theta}^{arg}$), normalized by number of arguments to sample. 
Therefore, with the learnable context representation $s_{\phi}(x)$ of input $x$, the end-to-end objective is to jointly learn $\{\theta,\phi\}$ that maximises the expected reward $R(x, a)\in \{-1,+1\}$ over the sampled actions ($a$), based on exact match with the gold answer. 

To mitigate the learning instability in such sparse confounding reward settings, we intialize with a simpler iterative \emph{hard-Expectation Maximization (EM)} learning objective, called Iterative Maximal Likelihood (IML) \citep{liang2017neural}. With the assumption that the sampled actions are extensive enough to contain the gold answer, IML greedily searches for the \emph{good} actions by fixing the policy parameters, and then maximises the likelihood of the \emph{best} action that led to the highest reward. We define \emph{good} actions ($\mathcal{A}^{good}$) as those that result in the gold answer itself and take a  conservative approach of defining \emph{best} among them as simply the most likely one according to the current policy. 
\begin{dmath}
J^{IML}(\theta,\phi) = \sum_x \max_{a\in \mathcal{A}^{good}} \log {P_{\theta,\phi} (a|x)} \label{eq:j_iml}
\end{dmath}
After the IML initialization, we switch to REINFORCE as the learning objective after a few epochs, where the goal is to maximise the expected reward ($J^{RL}(\theta,\phi) = \sum_x \E_{P_{\theta,\phi}(a|x)} R(x, a)$) as
\begin{dmath}
\nabla_{(\theta,\phi)} J^{RL} = \sum_x \sum_{a\in\mathcal{A}} P_{\theta,\phi}(a|x)  (R(x, a) - B(x)) \nabla_{\theta,\phi} (\log P_{\theta,\phi}(a|x))\label{eq:j_rl}
\end{dmath}
where $B(x)$ is simply the average (baseline) reward obtained by the policy for that instance $x$.
Further, in order to mitigate overfitting, in addition to $L_2$-regularization and dropout, we also add entropy based regularization over the argument sampling distribution, in each of the sampling networks.

\section{Experiments}

We now empirically compare the \emph{exact-match} performance of {\shortname} with SoTA baselines on versions of DROP dataset and also examine how it fares in comparison to strong supervised skylines.
\\
The \textbf{Primary Baselines} for {\shortname} are the explicit reasoning based \textbf{NMN} \citep{Gupta2020Neural}  which uses additional strong supervision and the BERT based language model \textbf{GenBERT} \citep{Geva2020InjectingNR} that does not embody any reasoning and autoregressively generates numeric answer tokens.
As the \textbf{Primary Dataset} we use \textbf{DROP-\emph{num}}, the subset of DROP with numerical answers. This subset contains 45K and 5.8K instances respectively from the standard DROP train and development sets. Originally, NMN was showcased on a very specific subset of DROP, restricted to the 6  reasoning-types it could handle, out of which three (\emph{count}, \emph{date-difference}, \emph{extract-number}) have numeric answers. This subset comprises 20K training and 1.8K development instances, out of which only 10K and 800 instances respectively have numerical answers. We further evaluate on this numerical subset, referred to as \textbf{DROP-Pruned-\emph{num}}. In both the cases, the training data was randomly split into 70\%:30\% for train and internal validation and the standard DROP development set was treated as the Test set. 

\begin{wrapfigure}[12]{r}{0.3\textwidth}
\includegraphics[width=\linewidth]{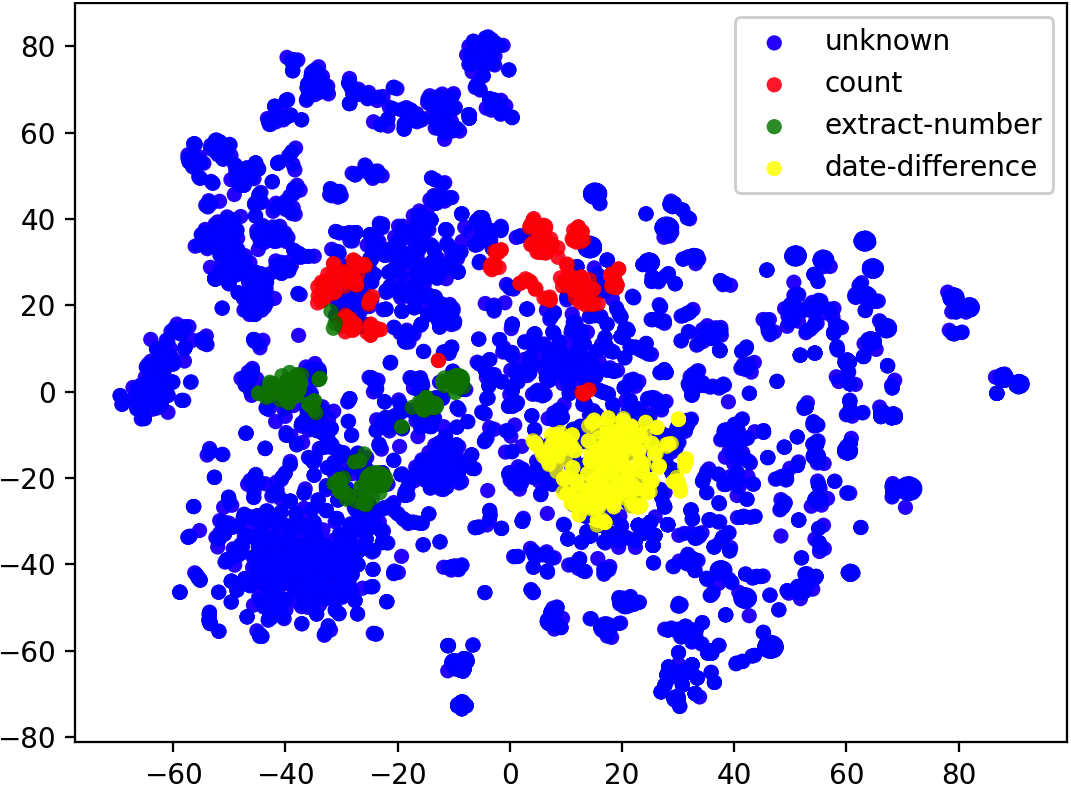}
\caption{t-SNE plot of DROP-\emph{num}-Test questions.}
\label{fig:tsne}
\end{wrapfigure}
\Cref{fig:tsne} shows the t-SNE plot of pretrained Sentence-BERT \citep{reimers-2019-sentence-bert} encoding of \emph{all} questions in DROP-\emph{num}-Test and also the DROP-Pruned-\emph{num}-Test subset with different colors (red, green, yellow) representing different types. Not only are the DROP-\emph{num} questions more diverse than the carefully chosen  DROP-Pruned-\emph{num} subset, the latter also forms well-separated clusters corresponding to the three reasoning types. Additionally, the average perplexity (using \texttt{nltk}) of the DROP-Pruned-\emph{num} and DROP-\emph{num} questions was found to be \emph{3.9} and \emph{10.65} respectively, further indicating the comparatively open-ended nature of the former.


For the primary baselines NMN and GenBERT, we report the performance on in-house trained models on the respective datasets, using the code open-sourced by the authors. The remaining results, taken from \cite{Geva2020InjectingNR}, \cite{nabert}, and \cite{ran-etal-2019-numnet}; refer to models trained on the full DROP dataset. All models use the same pretrained BERT-base. 
Also note that a primary requirement of all models other than GenBERT and {\shortname} \ie for NMN, MTMSN, NABERT, NAQANET, NumNet, is the exhaustive enumeration of the output space of all possible discrete operations. This simplifies the QA task to a classification setting, thus alleviating the need for discrete reasoning in the inference processs. 

\begin{wraptable}{r}{0.27\textwidth}
\caption{DROP-\emph{num}-Test Performance of Baselines and {\shortname}.} 
\centering
\scalebox{0.8}{
{\setlength{\tabcolsep}{0.25em}
          \begin{tabular}{c c c c}\\ 
          \multicolumn{3}{l}{\textbf{Supervision Type}} & \textbf{Acc.} (\%) \\ 
         \textbf{Prog.} & \textbf{Exec.} & \textbf{QAtt.} & \\\midrule
         \multicolumn{4}{c}{\textbf{NMN-\emph{num} variants}} \\ 
         \green{\xmark} & \red{\cmark} & \red{\cmark} & 11.77\\ 
         \red{\cmark} & \green{\xmark} & \red{\cmark} & 17.52\\
         \red{\cmark} & \red{\cmark} & \green{\xmark} & 18.27\\ 
         \red{\cmark} & \green{\xmark} & \green{\xmark} & 18.54\\
         \green{\xmark} & \red{\cmark} & \green{\xmark} & 12.27\\
         \green{\xmark} & \green{\xmark} & \red{\cmark} & 11.80\\
         \green{\xmark} & \green{\xmark} & \green{\xmark} & 11.70\\
         \midrule
         \multicolumn{4}{c}{\textbf{GenBERT}}  \\ 
         \green{\xmark} & \green{\xmark} & \green{\xmark} & 42.30\\ 
         \multicolumn{4}{c}{\textbf{GenBERT-\emph{num}}} \\  
         \green{\xmark} & \green{\xmark} & \green{\xmark} & 38.41\\
         \midrule
         \multicolumn{4}{c}{\textbf{\shortname}} \\ 
         \green{\xmark} & \green{\xmark} & \green{\xmark} & \textbf{50.97}\\\bottomrule
          \end{tabular}}}
    \label{tab:res_drop}
\end{wraptable}

Table \ref{tab:res_drop} presents our primary results on \textbf{DROP-\emph{num}}, comparing the performance of {\shortname} (accuracy of the top-1 sampled action by the RL agent) with various ablations of NMN (provided in the authors' implementation) by removing atleast one of \textbf{Prog}ram, \textbf{Exec}ution, and \textbf{Q}uery \textbf{Att}ention supervision (\Cref{appendix:nmn_limitations}) and GenBERT models with pretrained BERT that are finetuned on DROP or DROP-\emph{num} (denoted as GenBERT and GenBERT-\emph{num}). For a fair comparison with our weakly supervised model, we do not treat NMN with all forms of supervision or GenBERT model pretrained with additional \emph{synthetic} numerical and textual data as comparable baselines. Note that these GenBERT variants indeed enjoy strong reasoning supervision in terms of gold arithmetic expressions provided in these auxiliary datasets. 

NMN's performance is abysmally poor, indeed a drastic degradation in comparison to its performance on the pruned DROP subset reported by \cite{Gupta2020Neural} and in our subsequent experiments in \Cref{tab:res_drop_small}. This can be attributed to their limitation in handling more diverse classes of reasoning and  open-ended queries in DROP-\emph{num}, further exacerbated by the lack of one or more types of strong supervision.\footnote{Both the results and limitations of  NMN in Table\ref{tab:res_drop} and \ref{tab:res_drop_small} were confirmed by the authors of NMN as well.} Our earlier analysis on the complexity of the questions in the subset and full DROP-\emph{num} further quantify the relative difficulty level of the latter. On the other hand, GenBERT delivers a mediocre performance, while  GenBERT-\emph{num} degrades additionally by 4\%, as learning from numerical answers alone further curbs the language modeling ability. Our model  performs significantly better than both these baselines, surpassing GenBERT by 8\% and the NMN baseline by around 32\%. This showcases the significance of incorporating explicit reasoning in neural models in comparison to the vanila large scale LMs like GenBERT. It also establishes the generalizability of such reasoning based models to more open-ended forms of QA, in comparison to contemporary modular networks like NMN, owing to its ability to handle both learnable and discrete modules in an end-to-end manner. 

Next, in Table \ref{tab:res_drop_small}, we compare the performance of the proposed {\shortname} with the same NMN variants (as in Table \ref{tab:res_drop}) on \textbf{DROP-Pruned-\emph{num}}. Some of the salient observations are: 
\begin{wraptable}{r}{0.45\textwidth}
\caption{DROP-Pruned-\emph{num}-Test Performance of  NMN variants and {\shortname}}. 
\centering
\scalebox{0.79}{
{\setlength{\tabcolsep}{0.6em}
    \begin{tabular}{ccc c lll}\toprule 
        \multicolumn{3}{c}{\textbf{Supervision-Type}} & \multirow{2}{0.3cm}{\textbf{Acc.} (\%)} & 
        \multirow{2}{0.5cm}{\textbf{\small{Count}}} & \multirow{2}{0.8cm}{\textbf{\small{Extract-num}}} & \multirow{2}{0.65cm}{\textbf{\small{Date-differ}}}\\
        \textbf{Prog.} & \textbf{Exec.} & \textbf{QAtt.} & \\ [+0.2em]
          \midrule
        \multicolumn{3}{c}{\textbf{NMN-\emph{num} Variants}} &  \\ 
          \red{\cmark} & \red{\cmark} & \red{\cmark} & \textbf{68.6} & 50.0 & \textbf{88.4} & 72.5\\ \hline
          \green{\xmark} &\red{\cmark}  &\red{\cmark}  & 42.4 & 24.1 & 73.9 & 36.4\\ 
          \red{\cmark} & \green{\xmark} &\red{\cmark}  & 54.3 &47.9 & 80.7 & 40.9\\ 
          \red{\cmark} & \red{\cmark}  & \green{\xmark} & 63.3 &45.5 & 81.1 &68.7\\ 
          \green{\xmark} & \green{\xmark} & \red{\cmark} & 48.2 & 38.1 & 72.4 & 41.9\\ 
          \red{\cmark} & \green{\xmark} & \green{\xmark} & 61.0 & 44.7 & 81.1 & 63.2\\ 
          \green{\xmark} & \red{\cmark} & \green{\xmark} & 62.3 & 43.7 & 84.1 & 67.7\\ 
          \green{\xmark} & \green{\xmark} & \green{\xmark} & 62.1 & 46.8 & 83.6 & 66.1\\ \bottomrule
         \multicolumn{3}{c}{\textbf{\shortname}} \\ 
         \green{\xmark} & \green{\xmark} & \green{\xmark} & 66.5 & \textbf{58.8} & 66.8 & \textbf{75.2}\\ \bottomrule
        \end{tabular}}}
    \label{tab:res_drop_small}
\end{wraptable}
  \Ni {\shortname} in fact reaches a performance quite close to the \emph{strongly supervised} NMN variant (first row), and is able to attain at least an improvement margin of $4\%$ over all other variants obtained by removing one or more types of supervision. This is despite all variants of NMN \emph{additionally} enjoying the exhaustive precompution of the output space of possible numerical answers; \Nii {\shortname} suffers only in the case of \emph{extract-number} type operations (\eg \emph{max,min}) that involve a more complex process of sampling arbitrary number of arguments \Niii  Performance drop of NMN is not very large when all or none of the strong supervision is present, possibly because of the limited diversity over reasoning types and query language; 
and \Niv Query-Attention supervision infact adversely affects NMN's performance, in absence of the \emph{program} and \emph{execution} supervision or both, possibly owing to an undesirable biasing effect. However when both supervisions are available, query-attention is able to improve the model performance by $5\%$. Further, we believe the test set of 800 instances is too small to get an unbiased reflection of the model's performances. 

In Table \ref{tab:res_drop2}, we additionally inspect recall over the top-$k$ actions sampled by {\shortname} to estimate how it fares in comparison to the strongly supervised skylines:
\Ni NMN with all forms of strong supervision; 
\Nii GenBERT variants +ND, +TD and +ND+TD further pretrained on synthetic \textbf{N}umerical and \textbf{T}extual \textbf{D}ata and both;
 \Niii reasoning-free hybrid models like {MTMSN \citep{hu2019multi} } and NumNet \citep{ran-etal-2019-numnet}, NAQANet \citep{Dua2019DROP} and NABERT, NABERT+ \citep{nabert}.
Note that both NumNet and NAQANet do not use pretrained BERT.
MTMSN achieves SoTA performance 
through a supervised framework of  training specialized predictors for each reasoning type to predict the numerical expression directly instead of learning to reason. While top-1 performance of {\shortname} (in Table \ref{tab:res_drop})
is $4\%$ worser than NABERT, Recall@top-2 is equivalent to the strongly supervised NMN, top-5 and top-10 is comparable to NABERT+, NumNet and  GenBERT models +ND, +TD and top-20 nearly achieves SoTA. Such promising recall over the top-$k$ actions suggests that more sophisticated RL algorithms with better exploration strategies can possibly bridge this performance gap.

\begin{wraptable}[15]{r}{0.35\textwidth}
\vspace{-2em}
\caption{Skylines \& {\shortname} top-$k$ performance on DROP-\emph{num}-Test}

    \centering
         \scalebox{0.8}{\begin{tabular}{lc} \toprule 
         \textbf{Strongly Supervised Models} & \textbf{Acc. (\%)}\\ \midrule
         NMN-\emph{num} (all supervision) & 58.10\\ \midrule 
         GenBERT+ND & 69.20\\ 
         GenBERT+TD & 70.50\\ 
         GenBERT+ND+TD & 75.20\\ \midrule
         NAQANet & 44.97\\ 
         NABERT & 54.27\\ 
         NABERT+ & 66.60\\ 
         NumNet & 69.74 \\ \midrule
         MTMSN & 75.00\\
         \bottomrule 
        \end{tabular}}
\scalebox{0.8}{
{\setlength{\tabcolsep}{0.2em}
        \begin{tabular}{c|c|c|c|c|c} \toprule 
         \multicolumn{6}{c}{\textbf{Recall@top-$k$ actions of {\shortname} (\%)}} \\ 
        $k=2$ & $k=3$ & $k=4$ & $k=5$ & $k=10$ & $k=20$\\ 
        58.6 & 63.0 & 65.4 & 67.4 & 72.3 & 74.2 \\ \bottomrule 
         
         \end{tabular}}}
    \label{tab:res_drop2}
\end{wraptable}

\section{Analysis \& Future Work}


\paragraph{Performance Analysis} Despite the notorious  instabilities of RL due to high variance, the training trend, as shown in Figure 5(a) is not afflicted by catastrophic forgetting. The sudden performance jump between epochs 10-15 is because of switching from  iterative ML initialization to REINFORCE  objective. 
Figure 5(b) shows the individual module-wise performance evaluated using the noisy pseudo-rewards, that indicate whether the action sampled by this module \emph{led} to the correct answer or not (details in  \Cref{appendix:RL}). Further, by bucketing the performance by the total number of passage entities in Figure  5(c), we observe that {\shortname} remains unimpacted by the increasing number of date/numbers, despite the action space explosion. On the other hand, GenBERT's performance drops linearly beyond 25 passage entities and NMN-\emph{num} degrades exponentially from the beginning, owing to its  direct dependency on the exponentially growing exhaustively precomputed output space.

\begin{table}[!htb]
\begin{minipage}{0.33\linewidth}
 \includegraphics[width=\linewidth,height=9em]{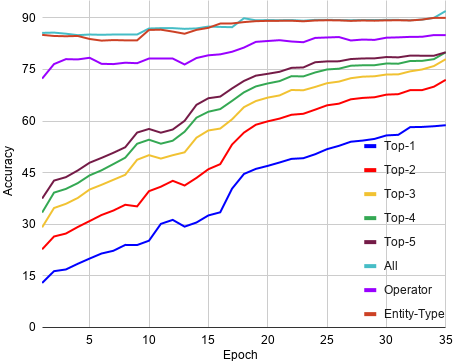}
\end{minipage}
\begin{minipage}{0.33\linewidth}
{\small
\scalebox{0.85}{
{\setlength{\tabcolsep}{0.25em}\begin{tabular}{ll}
\toprule
\textbf{Module} & \textbf{Performance} \\
\midrule
     Sample 1 Argument & 54\% (Acc.)\\ 
     Sample 2 Argument & 52\% \ditto\\ 
     Counter & 50\% \ditto\\ 
     Entity Ranker & 53\% \ditto\\ 
     Operator Predictor & 78\% \ditto\\ 
     Entity Type Predictor & 83\% \ditto\\ 
     \midrule
     Overall Action Sampler  & 84\% (Rec@All)\\
     \bottomrule
\end{tabular}}}}
\end{minipage}
\begin{minipage}{0.33\linewidth}
 \includegraphics[width=\linewidth,height=9em]{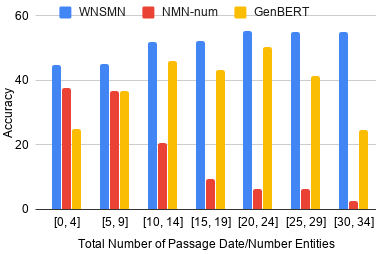}
\end{minipage}
\label{tab:res_module}
\caption*{Figure 5: (a) Training trend showing the Recall@top-$k$ and all actions, accuracy of Operator and Entity-type Predictor, estimated based on noisy psuedo rewards (\Cref{appendix:RL}), (b) Module-wise performance (using pseudo-reward) on DROP-\emph{num}-Test, (c) Bucketing performance by total number of passage entities for \shortname, and the best performing NMN and GenBERT model from Table \ref{tab:res_drop}.}
\end{table}
\paragraph{More Stable RL Framework} The training trend in Figure 5(a) shows early saturation and the module-wise performance indicates overfitting despite the regularization tricks in \Cref{subsec:RL} and \Cref{appendix:RL}. While more stable RL algorithms like Actor-Critic, Trust Region Policy Optimization \citep{pmlr-v37-schulman15}
or Memory Augmented Policy Optimization \citep{NIPS2018_8204}
can mitigate these issues, we leave them for future exploration. Also, though this work's objective was to train module networks with weak supervision, the sparse confounding rewards in the exponential action space indeed render the RL training quite challenging. One practical future direction to bridge the performance gap would be to pretrain with strong supervision on at least a subset of reasoning categories or on more constrained forms of synthetic questions, similar to GenBERT. Such a setting would require inspection and evaluation of generalizability of the RL model to unknown reasoning types or more open-ended questions. 

\section{Related Work}
In this section we briefly compare our proposed {\shortname} to the two closest genre of models that have proven quite successful on DROP \footnote{A more detailed related work section is presented in the Appendix \ref{appendix:related_work}} i) reasoning free hybrid models NumNet, NAQANet, NABERT, NABERT+, MTMSN, and NeRd ii) modular network for reasoning NMN. Their main distinction with {\shortname} is that in order to address the challenges of weak supervision, they obtain program annotation from the QA pairs through i) various heuristic parsing of the templatized queries in DROP to get supervision of the reasoning type (max/min, diff/sum, count, negate). ii) exhaustive search over all possible discrete operations to get supervision of the arguments in the reasoning.

Such heuristic supervision makes the learning problem significantly simpler in the following ways  
\begin{itemize}[leftmargin=*]
    \item These models enjoy supervision of specialized program that have explicit information of the type of reasoning to apply for a question \eg SUM(10,12)
    \item A simplistic (contextual BERT-like) \emph{reader} model to read query related information from the passage trained with direct supervision of the query span arguments at each step of the program
    \item A \emph{programmer} model that can be directly trained to decode the specialized programs 
    \item  \emph{Executing} numerical functions (\eg \emph{difference, count, max, min}) either by i) training purely neural modules in a strong supervised setting using the annotated programs or by ii) performing the actual discrete operation as a post processing step on the model's predicted program. For each of these previous works, it is possible to directly apply the learning objective on the space of decoded program, without having to deal with the discrete answer or any non-differentiability. 
\end{itemize}

However, such heuristic techniques of program annotation or exhaustive search is not practical as the language of questions or the space of discrete operations become more complex. Hence {\shortname} learns in the challenging weak-supervised setting without any additional annotation through
\begin{itemize}[leftmargin=*]
    \item A noisy symbolic query decomposition that is oblivious to the reasoning type and simply based on generic text parsing techniques 
    \item An entity specific cross attention model extracting passage information relevant to each step of the decomposed query and learning an attention distribution over the entities of each type
    \item Learning to apply discrete reasoning by employing neural modules that learn to sample the operation and the  entity arguments 
    \item Leveraging a combination of neural and discrete modules when executing the discrete operation, instead of using only neural modules which need strong supervision of the programs for learning the functionality
    \item Fundamentally different learning strategy by incorporating inductive bias through auxiliary losses and Iterative Maximal Likelihood for a more conservative initialization followed by REINFORCE
\end{itemize}

These reasoning-free hybrid models are not comparable with {\shortname} because of their inability to learn in absence of any heuristic program annotation. Instead of learning to reason based on only the final answer supervision, they reduce the task to learning to decode the program, based on heuristic program annotation. NMN is the only reasoning based model that employ various auxiliary losses to learn even in absence of any additional supervision, similar to us. 

To our knowledge {\shortname} is the first work on modular networks for fuzzy reasoning over text in RC framework, to handle the challenging cold start problem of the weak supervised setting without needing any additional specialized supervision of heuristic programs.

\section{Conclusion}
In this work, we presented \longname\ for numerical reasoning based MRC based on a generalized framework of query parsing to noisy heuristic programs. It trains both neural and discrete reasoning modules end-to-end in a Deep RL framework with only discrete reward based on exact answer match. Our empirical analysis on the \emph{numerical-answer only subset} of DROP showcases significant performance improvement of the proposed model over SoTA NMNs and Transformer based language model GenBERT, when trained in comparable weakly supervised settings. While, to our knowledge, this is the first effort towards training modular networks for fuzzy reasoning over RC in a weakly-supervised setting, there is significant scope of improvement, such as employing more sophisticated RL framework or by leveraging the pretraining of reasoning.
\nocite{NIPS2018_8204}
\nocite{pmlr-v37-schulman15}
\nocite{nmninterpret:acl20}

\bibliography{iclr2021_conference}
\bibliographystyle{iclr2021_conference}
\newpage
\appendix
\section{Appendix}
\subsection{Qualitative Analysis}
\begin{table}[H]
\begin{center}
{\scriptsize
\begin{tabular}{|p{9cm}|p{4cm}|}\hline
\textbf{\longname} & \textbf{GenBERT} \\ \hline
\multicolumn{2}{|p{13cm}|}{\vspace{0.02cm}\textbf{1.\hspace{0.5em}Query:} how many times did a game between the patriots versus colts result in the exact same scores?, \textbf{Ans:} 2}\\
\multicolumn{2}{|p{13cm}|}{\hspace{1.3em}\textbf{Num. of Passage Entities}: Date(10), Number(9)}\\ \hline 
\vspace{0.01cm}
D, N = \emph{Entity-Attention}(`how many times') // \emph{D, N are the attention distribution over date and number entities} & \vspace{0.01cm}Predicted AnsType:  Decoded \\
D1, N1 = \emph{Entity-Attention}(`did a game between the patriots versus colts result in the exact same scores', (D, N)) & Decoder output: 2\\
`Number', `Count' = \emph{EntType-Operator-Selector}(`how many times', Query) & Span extracted: ``colts''\\ 
\textbf{Answer} 2 = \emph{Count}(N1) & \textbf{Answer} = 2\\  \hline 

\multicolumn{2}{|p{13cm}|}{\vspace{0.02cm}\textbf{2.\hspace{0.5em}Query:} how many people in chennai, in terms of percent population, are not hindu?, \textbf{Ans:} 19.3}\\
\multicolumn{2}{|p{13cm}|}{\hspace{1.3em}\textbf{Num. of Passage Entities}: Date(2), Number(26)}\\ \hline 
\vspace{0.01cm}D, N = \emph{Entity-Attention}(`how many people in chennai, in terms of percent population') &  \vspace{0.01cm}Predicted AnsType: Decoded\\ 
D1, N1 = \emph{Entity-Attention}(`are not hindu', (D, N)) & Decoder output: 19.3\\
`Number', `Negate' = \emph{EntType-Operator-Selector}('are not hindu', Query) & Span extracted: ``80.7'' \\
1 = Count(N) & \textbf{Answer} = 19.3\\ 
\{80.7\} = \emph{Sample-Arbitrary-Arguments}(N1, 1) &\\ 
\textbf{Answer} = 19.3 = \emph{Negate}(\{80.7\}) & \\\hline

\multicolumn{2}{|p{13cm}|}{\vspace{0.02cm}\textbf{3.\hspace{0.5em}Query:} how many more percent of the population was male than female?, \textbf{Ans:} 0.4}\\
\multicolumn{2}{|p{13cm}|}{\hspace{1.3em}\textbf{Num. of Passage Entities}: Date(4), Number(29)}\\ \hline
\vspace{0.01cm}
D, N = \emph{Entity-Attention}(`how many')& \vspace{0.01cm} Predicted AnsType: Decoded \\
D1, N1 = \emph{Entity-Attention}(`more percent of the population was male', (D, N)) & Decoder output: 3.2 \\ 
D2, N2 = \emph{Entity-Attention}(`than female', (D, N)) & Span extracted: ``49.8'' \\
`Number',`Difference' = \emph{EntType-Operator-Selector}('how many', Query) & \textbf{Answer} = 3.2 \\
50.2 = Sample-1-Argument(N1) & \\
49.8 = \emph{Sample-2-Argument}(N2) & \\
\textbf{Answer} = 0.4 = \emph{Difference}(\{50.2, 49.8\}) & \\ \hline

\multicolumn{2}{|p{13cm}|}{\vspace{0.02cm}\textbf{4.\hspace{0.5em}Query:} how many more, in percent population of aigle were between 0 and 9 years old than are 90 and older?, \textbf{Ans:} 9.8}\\ 
\multicolumn{2}{|p{13cm}|}{\hspace{1.3em}\textbf{Num. of Passage Entities}: Date(0), Number(25)}\\ \hline 
\vspace{0.01cm}
D, N = \emph{Entity-Attention}(`how many more') & \vspace{0.01cm}Predicted AnsType: Decoded\\
D1, N1 = \emph{Entity-Attention}(`in percent population of aigle were between 0 and 9 years old', (D, N)) & Decoder output: 1.7 \\
D2, N2 = \emph{Entity-Attention}(`than are 90 and older', (D, N)) & Span extracted: ``0.9''\\
`Number', `Difference' = \emph{EntType-Operator-Selector}(`how many more', Query) & \textbf{Answer} = 1.7\\
10.7 = \emph{Sample-1-Argument}(N1) & \\
0.9 = \emph{Sample-1-Argument}(N2) & \\
\textbf{Answer} = 9.8 = \emph{Difference}(\{10.7, 0.9\}) & \\\hline 

\multicolumn{2}{|p{13cm}|}{\vspace{0.02cm}\textbf{5.\hspace{0.5em}Query:} going into the 1994 playoffs, how many years had it been since the suns had last reached the playoffs?, \textbf{Ans:} 3}\\ 
\multicolumn{2}{|p{13cm}|}{\hspace{1.3em}\textbf{Num. of Passage Entities}: Date(3), Number(17)}\\ \hline 
\vspace{0.01cm}
D, N = \emph{Entity-Attention}(`going into the 1994 playoffs : how many years') & \vspace{0.01cm}Predicted AnsType: Decoded \\ 
D1, N1 = \emph{Entity-Attention}(`had it been since the suns had last reached the playoffs', (D, N)) & Decoder output: 7\\
`Date', `Difference' = \emph{EntType-Operator-Selector}(`going into the 1994 playoffs : how many years', Query) & Span extracted:``1991'' \\
\{1991, 1994\} = \emph{Sample-2-Argument}(D) & \textbf{Answer} = 7\\
\textbf{Answer} = 3 = \emph{Difference}(\{1991, 1994\}) & \\\hline 

\multicolumn{2}{|p{13cm}|}{\vspace{0.02cm}\textbf{6.\hspace{0.5em}Query:} how many more points did the cats have in the fifth game of the AA championship playoffs compared to  st. paul saints?,}\\
\multicolumn{2}{|p{13cm}|}{\hspace{1.3em}\textbf{Ans:} 3}\\
\multicolumn{2}{|p{13cm}|}{\hspace{1.3em}\textbf{Num. of Passage Entities}: Date(3), Number(12)}\\ \hline 
\vspace{0.01cm}
D, N = \emph{Entity-Attention}(`how many') & \vspace{0.01cm}Predicted AnsType:  Decoded\\ 
D1, N1 = \emph{Entity-Attention}(`more points did the cats have in the fifth game of the AA championship playoffs', (D, N)) & Decoder output: 3\\
D2, N2 = \emph{Entity-Attention}(`compared to the st. paul saints', (D, N)) & Span extracted: ``4 - 1 in the fifth game''\\
`Number', `Difference' = \emph{EntType-Operator-Selector}(`how many', Query) & \textbf{Answer} = 3\\
5.0 = \emph{Sample-1-Argument}(N1) &\\
2.0 = \emph{Sample-1-Argument}(N2) &\\
\textbf{Answer} = 3.0 = \emph{Difference}(\{5.0, 2.0\}) &\\ \hline 

\multicolumn{2}{|p{13cm}|}{\vspace{0.02cm}\textbf{7.\hspace{0.5em}Query:} how many total troops were there in the battle?, \textbf{Ans:} 40000}\\
\multicolumn{2}{|p{13cm}|}{\hspace{1.3em}\textbf{Num. of Passage Entities}: Date(1), Number(3)}\\ \hline 
\vspace{0.01cm}
D, N = \emph{Entity-Attention}(`how many total troops') & \vspace{0.01cm}Predicted AnsType: Decoded\\
D1, N1 = \emph{Entity-Attention}(`were there in the battle', (D, N)) & Decoder output: 100000\\
`Number', `Sum' = \emph{EntType-Operator-Selector}('how many total troops', Query) & \\
2 = \emph{Count}(N1) & Span extracted: ``10000 korean troops''\\
\{10000.0, 30000.0\} = \emph{Sample-Arbitrary-Arguments}(N1, 2) & \textbf{Answer} = 100000\\
\textbf{Answer} = 40000.0 = \emph{Sum}(\{10000.0, 30000.0\}) & \\ \hline 
\end{tabular}
}
\end{center}
\label{tab:examples1}
\end{table}

\begin{table}[!htb]
\begin{center}
{\scriptsize
\begin{tabular}{|p{5.8cm}|p{3.6cm}|p{3.3cm}|}\hline
\textbf{\longname} & \textbf{NMN-\emph{num}} & \textbf{GenBERT} \\ \hline 
\multicolumn{3}{|p{13cm}|}{\vspace{0.02cm}\textbf{8.\hspace{0.5em}Query:} how many field goals did sebastian janikowski and kris brown both score each? \textbf{Ans:} 2}\\
\multicolumn{3}{|p{13cm}|}{\hspace{1.3em}\textbf{Num. of Passage Entities}: Date(0), Number(9)}\\ \hline
\vspace{0.01cm}
D, N = \emph{Entity-Attention}(`how many field goals')
 & \vspace{0.01cm}P1 = \emph{Find-Passage-Attention}() & \vspace{0.01cm}Predicted AnsType: Decoded \\ 
D1, N1 = \emph{Entity-Attention}(`did sebastian janikowski and kris brown both score each', (D, N)) & P2 = \emph{Filter-Passage-Attention}(P1) & Decoder output: 2 \\ 
`Number', `Count' = \emph{EntType-Operator-Selector}(`how many field goals', Query) & 2 = \emph{Passage-Attn-To-Count}(P2) & Span extracted: ``33 - yard'' \\
\textbf{Answer} = 2.0 = \emph{Count}(N1) & \textbf{Answer} = 2 & \textbf{Answer} = 2\\ \hline

\multicolumn{3}{|p{13cm}|}{\vspace{0.02cm}\textbf{9.\hspace{0.5em}Query:} how many years was between the oil crisis and the energy crisis? \textbf{Ans:} 6}\\
\multicolumn{3}{|p{13cm}|}{\hspace{1.3em}\textbf{Num. of Passage Entities}: Date(19), Number(14)}\\ \hline
\vspace{0.01cm}D1, N1 = \emph{Entity-Attention}(`was between the oil crisis and the energy crisis') & \vspace{0.01cm}\emph{year-diffs} $\in \mathbbm{R}^{40}$ (// generated exhaustive output space of all differences)& \vspace{0.01cm}Predicted AnsType: Decoded\\
D, N = \emph{Entity-Attention}(`how many years', D1, N1) & P1 = \emph{Find-Passage-Attention}() & Decoder output: 3\\
`Date', `Difference' = \emph{EntType-Operator-Selector}(`how many years', Query) & 6 = \emph{Year-Difference}(P1, \emph{year-diffs}) & Span extracted: ``1973''\\
\{1973, 1979\} = \emph{Sample-2-Argument}(D) & \textbf{Answer} = 6.0 & \textbf{Answer} = 3\\
\textbf{Answer} = 6.0 = \emph{Difference}(\{1973, 1979\}) & &\\ \hline 

\multicolumn{3}{|p{13cm}|}{\vspace{0.02cm}\textbf{10.\hspace{0.5em}Query:} how many yards was the longest touchdown pass? \textbf{Ans:} 40}\\
\multicolumn{3}{|p{13cm}|}{\hspace{1.3em}\textbf{Num. of Passage Entities}: Date(0), Number(5)}\\ \hline
\vspace{0.01cm}
D, N = \emph{Entity-Attention}(`how many yards was the') &\vspace{0.01cm} P1 = \emph{Find-Passage-Attention}() &\vspace{0.01cm} Predicted AnsType: Extract-Span\\
D1, N1 = \emph{Entity-Attention}(`longest touchdown pass', (D, N)) & N1 = \emph{Find-Passage-Number}(P1)  & Decoder output: 43 \\ 
`Number', `Sum' = \emph{EntType-Operator-Selector}(`how many yards was the', Query) & 40 = \emph{Find-Max-Num}(N1) &  Span extracted: ``40''\\ 
1 = \emph{Count}(N) & \textbf{Answer} = 40 & \textbf{Answer} = 40\\
\{40.0\} = \emph{Sample-Arbitrary-Argument}(N, 1) & &\\
\textbf{Answer} = 40.0 = \emph{Sum}(\{40.0\}) & & \\\hline 
\end{tabular}
}
\caption{Example questions from DROP-\emph{num} along with predictions of the Proposed model {\shortname} and the best performing versions of the NMN-\emph{num} and GenBERT baselines from \Cref{tab:res_drop}. Detailed elaborations of outputs of these three models below: \vspace{0.5em}\\
(i) \textbf{{\shortname}} first parses the dependency structure in the query into a program-form. Next, for each step of the program, it generates an attention distribution over the date and number entities. \emph{Entity-Attention} refers to that learnt entity-specific cross attention described in \Cref{subsec:extraction}. It then performs the discrete reasoning by sampling an operation and specific entity-arguments, in order to reach the answer.  \emph{EntType-Operator-Selector} refers to the Entity-Type and Operator Predictor in Operator Sampling Network and Sample-*-Argument refers to the Argument Sampling Network described in \Cref{subsec:select}. \emph{Sum/Difference/Logical-Not} are some of the discrete operations that are executed to get the answer. In some of the cases, (\eg Query 3.) despite wrong parsing the model was able to predict the correct operation even though the root clause did not have sufficient information. In Query 10., the correct operation is \emph{Max}, but {\shortname} reaches the right answer by sampling only the maximum number entity through the Sample-Arbitrary-Argument network and then applying a spurious Sum operation on it.\vspace{0.5em}\\ (ii) On the other hand, the steps of the program generated by \textbf{NMN-\emph{num}} first compute or further filter attention distribution over the passage or entities which are then fed into the learnable modules (\emph{Passage-Attn-To-Count}, \emph{Year-Difference}) that predict the answer. In order to do so, it needs to precompute all possible outputs of numerical operations that generate new numbers for e.g. \emph{year-diffs} in Example 9. Because of the relatively poorer performance of NMN-\emph{num}, its outputs are only reported for the last 3 instances, which were cherrypicked based on NMN-\emph{num}'s predictions.\vspace{0.5em}\\ (iii) \textbf{GenBERT} first predicts whether the answer should be decoded or extracted from passage span and accordingly uses the Decoder output or extracted span as the answer. By design, the modular networks provide a more interpretable output than the monolithic encoder-decoder model GenBERT. }
\end{center}
\label{tab:examples2}
\end{table}
\subsection{Implementation \& Pseudo-Code}
The source-code and models pertaining to this work would be open-sourced on acceptance of this work. A detailed pseudo-code of the {\shortname} algorithm is provided below.

\begin{algorithm}
\caption{{\shortname} Algorithm}
\begin{algorithmic}
\STATE \textbf{Input:} (Query ($q$), Passage ($p$)) = $x$
\STATE \textbf{Output (or Supervision)}: Answer($y$) $\in \mathbbm{R}$
\vspace{0.5em}
\STATE \textbf{Preprocessing:}\\
\bindent
\STATE [$num_1$, $num_2$, $\ldots$, $num_N$] = $Num$ = Extract-Numbers($p$) \hfill\COMMENT{\emph{{\color{blue}// Number and Date}}}
\STATE [$date_1$, $date_2$, $\ldots$, $date_D$] = $Date$ = Extract-Dates($p$)\hfill\COMMENT{\emph{{\color{blue}// Entity and Passage Mentions}}}
\eindent
\vspace{0.5em}
\STATE \textbf{Inference:}\\

\bindent
\STATE $[ (q_{1}, ref_{1}), \ldots (q_{k}, ref_{k}), \ldots (q_{l}, ref_{l})]$ = $Program$ = Query-Parsing($q$)
\vspace{0.5em}
\FOR{ step $(q_{k}, ref_{k}) \in Program$}
\STATE $(\mA^{num}_{k}$, $\mathcal{T}^{num}_k$), ($\mA^{date}_{k}$, $\mathcal{T}^{num}_k$) = Entity-Attention($q_k$, $p$, $ref_{k}$, $Num$, $Date$)\hfill\COMMENT{\emph{{\color{blue}\Cref{subsec:extraction}}}}
\ENDFOR
\vspace{0.5em}
\STATE $\gL^{num}_{aux}$, $\gL^{date}_{aux}$ = Entity-Inductive-Bias($\tA^{num}$, $\tA^{date}$)\hfill\COMMENT{\emph{{\color{blue}\Cref{eq:unsup-loss}}}}
\STATE $\gL_{aux} =  \gL^{num}_{aux} + \gL^{date}_{aux}$ 

\vspace{0.5em}

\STATE $q_{l}$ = \emph{Query Span} Argument of Last Step
\hfill\COMMENT {\emph{{\color{blue}// Program Arguments and Stacked Attention }}}
\STATE $ref_{l}$ = \emph{Reference} Argument of Last Step
\hfill\COMMENT {\emph{{\color{blue}// Map over Entities for Last Step}}}
\STATE $\mathcal{T}^{num}$ = \{${\mathcal{T}^{num}_k | k \in ref_{l}}$\}, $\mathcal{T}^{date}$ = \{${\mathcal{T}^{date}_k | k \in ref_{l}}$\}
\vspace{0.5em}
\STATE $Operators$ = \{$op_1$, $op_2$, \ldots , $op_{k1}$\} = Operator-Predictor($q_{l}$, $q$) \hfill\COMMENT {\emph{{\color{blue}// Operator and EntityType}}}
\STATE $EntTypes$ = \{$type_{1}$, $type_{2}$, \ldots, $type_{k1}$\} = Entity-Type-Predictor($q_{l}$, $q$)\hfill\COMMENT {\emph{{\color{blue}// Sampling}}}
\vspace{0.5em}
\STATE $Actions$ = \{\}\hfill\COMMENT {\emph{{\color{blue}// Action Sampling for each Operator}}}
\FOR { $op$, $type$ $\in$ ($Operators$, $EntTypes$)} 
\IF{$type$ \textbf{is} Number} 
\STATE $\mathcal{T} = \mathcal{T}^{num}$
\ELSIF{$type$ \textbf{is} Date}
\STATE $\mathcal{T} = \mathcal{T}^{date}$
\ENDIF
\IF{$op$ \textbf{is} \texttt{diff}} 
\IF{$|ref_l|== 2$}
\STATE $arg1$ = \{$arg1_1$, $arg1_2$, $\ldots$,  $arg1_{k2}$\} = Sample-1-Argument($\mathcal{T}_0$)
\STATE $arg2$ = \{$arg2_1$, $arg2_2$, $\ldots$,  $arg2_{k2}$\} = Sample-1-Argument($\mathcal{T}_1$)
\STATE $args$ = \{$(a1, a2) |\hspace{0.5em}(a1, a2) \in (arg1, arg2)$\}
\ELSIF{$|ref_l|== 1$}
\STATE $args$ = \{$arg_1$, $arg_2$, $\ldots$,  $arg_{k2}$\} =
  Sample-2-Argument($\mathcal{T}_0$)
\ENDIF 
\ELSIF{$op$ \textbf{is} \texttt{count}}
\STATE  $args$ = \{$count_1$, $count_2$, $\ldots$, $count_{k2}$\} =  Count-Network($\sum_{j}\mathcal{T}_j$)
\ELSE
\STATE  $args$ = \{$arg_1$, $arg_2$, $\ldots$,  $arg_{k2}$\} =  Sample-Arbitrary-Argument($\sum_{j}\mathcal{T}_j$)
\ENDIF
\STATE $probs$ = \{$(p^{type}*p^{op}*p) | p \in p^{arg}\} \in \mathbbm{R}^{k2}$ \hfill\COMMENT {\emph{{\color{blue}// $p$'s refer to the corresponding probabilities}}} 
\STATE $answers$ = \{Execute-Discrete-Operation($type$, $op$, $arg$)$ |\hspace{0.5em} arg \in args\} \in \mathbbm{R}^{k2}$
\STATE $actions$ = \{($prob$, $answer$)$| \hspace{0.5em}prob \in probs, answer \in answers\}$
\STATE $Actions$ = $Actions$ $\cup$ $actions$
\ENDFOR
\eindent
\vspace{0.5em}
\STATE \textbf{Training}:
\bindent
\FOR { $i \in \{1, \ldots, N_{IML} + N_{RL}\}$}
\FOR { $(x,y) \in \mathcal{D}$}
\STATE $\mathcal{A}(x) \longleftarrow Actions$ sampled for input($x$) \hfill\COMMENT {\emph{{\color{blue}// Using above Algorithm}}}
\STATE $R(x,a,y) \longleftarrow$ Exact Match Reward for action $a$ for instance $x$ with gold answer $y$
\IF{$i \le N_{IML}$}
\STATE $(\theta, \phi)\longleftarrow \displaystyle\max_{\theta,\phi} J^{IML}$ over ($\mathcal{A}, R) + \displaystyle\min_{\phi} \gL_{aux}$\hfill\COMMENT{\emph{{\color{blue}$J^{IML}$ from  \Cref{eq:j_iml}}}}
\ELSE
\STATE $(\theta, \phi)\longleftarrow \displaystyle\max_{\theta,\phi} J^{RL}$ over ($\mathcal{A}, R) + \displaystyle\min_{\phi} \gL_{aux}$\hfill\COMMENT{\emph{{\color{blue}$J^{RL}$ from \Cref{eq:j_rl}}}}
\ENDIF
\ENDFOR
\ENDFOR
\eindent
\label{appendix:algorithm}
\end{algorithmic}
\end{algorithm}

\subsection{Qualitative Inspection of {\shortname} Predictions}
\label{appendix:analysis}
\begin{table}[h]
{\small
 \begin{tabular}{|p{10cm}|p{3cm}|} \hline
\multicolumn{2}{|l|}{\textbf{Good Action:} Action Resulting in exact match with gold answer} \\ \hline
\multicolumn{2}{|l|}{\textbf{Correct Action:} Action Manually annotated to be correct} \\ \hline\hline

Number of test instances (DROP-num Test) & 5800 \\ \hline
Number of instances with atleast 1 good action & 4868 \\ \hline
Number of instances with more than 1 good action & 2533 \\ \hline\hline 

Average number of good actions (where there is atleast 1 good action) & 1.5  \\ \hline
Average number of good actions (where there is more than 1 good action) & 2.25 \\ \hline\hline

Number of instances where the top-1 action is good action & 2956 \\ \hline
Number of instances where top-1 is the  only good action & 2335 (79\% of 2956) \\ \hline
Number of instances with possibility of top-1 action being spuriously good & 620 (21\% of 2956) \\ \hline\hline

Number of instances manually annotated (out of possible cases of spurious top-1 action) & 334 (out of 620)  \\ \hline
Number of instances where top-1 action is found to be spurious & 28  (8.4\% of 334) \\ \hline
Avg Ratio of Probability of Top Action and Maximum Probability of all other spuriously good actions (if any) & 4.4e+11 \\ \hline
\end{tabular}
\caption{Analysis of the predictions of WNSMN on DROP-\emph{num} Test}
}
\end{table}

Generic Observations/Notes
\begin{itemize}
    \item Note: When the model selects a single number in the Argument Sampling network and the Operator sampled is not of type count, we forcefully consider the operation as a NO-OP. For example sum/min/max over a single number or date is treated as NO-OP.
    \item One potential source of spuriously correct answer is the neural `counter' module which can predict numbers in [1, 10]. However, out of the cases where atleast one of the top-50 actions is a good action we observe that the model is able to learn when the answer is directly present as an entity or can be obtained through (non count) operations over other entities and when it cannot be obtained directly from the passage but needs to aggregate (i.e., count) over multiple entities. Table \ref{appendix:example_analysis} below gives some examples of \emph{hard} instances where the {\shortname} Top-1 prediction was found to be correct.
\end{itemize}

\begin{table}[h]
{\small
\begin{tabular}{|p{5cm}|p{7cm}|p{1cm}|}\hline 
\textbf{True Reasoning} & \textbf{Model Prediction} & \textbf{Count} \\ \hline

\emph{negate} a passage entity \ie 100 - number & the model was able to select \emph{negate} of the correct entity as the top action. & 34\\ \hline 

\emph{min/max} of a set of passage entities &  the model instead directly sampled the correct minimum/maximum entity as a single argument and then applied NO-OP operation over it. & 11 \\ \hline 

\emph{select} one of the passage entities 
& the model was able to select the right entity and apply NO-OP on it as the top action. & 18 \\ \hline 

\emph{count} over passage entities 
& the model was able to put count as the top action and the spurious actions came much lower with almost epsilon probability & 88 \\ \hline 

\emph{difference} over passage entities (the same answer could be spuriously obtained by other non-difference operations over unrelated entites) & the model was able to put difference as the top action and the spurious actions came much lower with almost epsilon probability & 89 \\ \hline 

\emph{difference} over passage entities (the same answer could be spuriously obtained by difference over other unrelated entities) & the model was able to put difference over the correct arguments as the top action & 66\\ \hline 

\end{tabular}
\caption{Case Study of the 306 instances manually annotated as Correct out of 334 instances}
}
\end{table}

\begin{table}[h]
{\small
\begin{tabular}{|p{6cm}|p{6cm}|p{1cm}|} \hline
\textbf{True Reasoning} & \textbf{Model Prediction} & \textbf{Count} \\ \hline  
difference of dates/months & count over years & 4\\ \hline
sum(number1, count([number2])  & count over numbers & 1\\ \hline
difference between entities & sum over two arguments (both arguments wrong) & 1\\ \hline
difference between entities & difference over two arguments (both arguments wrong) & 1\\ \hline
difference between entities & count over entities & 1\\ \hline
difference between entities &  sum over arguments (one correct) (correct action was taken in one of the other top-5 beams)  & 2\\ \hline
question is vague/incomplete/could not be answered manually & count or difference & 2\\ \hline
counting over text spans (Very rare type of question, only 2 found out of 334) & wrong operator & 2\\ \hline
miscelleneous &  wrong operator  & 7\\ \hline
miscelleneous & correct operator wrong arguments (one correct) & 2\\ \hline
miscelleneous & correct operator wrong arguments (all wrong) & 5 \\ \hline
\end{tabular}
\caption{Case Study of the 28 instances manually annotated as Wrong out of 334 instances.}
}
\end{table} 

\begin{table}[h]
{\scriptsize
\begin{tabular}{|p{3cm}|p{4cm}|p{6cm}|} \hline 
\textbf{Question} & \textbf{Relevant Passage Excerpt} & \textbf{Model Prediction Analysis}\\ \hline 
How many printing had Green Mansions gone through by 1919?	& ``W. H. Hudson which went through nine printings by 1919 and sold over 20,000 copies.... '' &	Model was able to rank the operation sum([9.0]) highest. the \emph{count-number} operator had near-epsilon probability, indicating that indeed it did not find any indication of the answer being 9 by counting entities over the passage. This is despite the fact that most of the "how many" type questions need counting. \\ \hline

The Steelers finished their 1995 season having lost how many games difference to the number of games they had won? &	``In 1995, the Steelers overcame a 3-4 start (including a 20-16 upset loss to the expansion 1995 Jacksonville Jaguars season) to win eight of their final nine games and finished with an  record, the second-best in the AFC''. &  Model had to avoid distracting numbers (3,4) and (20,16) to understand that the correct operation is difference of (9-8) \\ \hline

How many more field goals did Longwell boot over Kasay?	&  ``26-yard field goal by kicker Ryan Longwell ... Carolina got a field goal with opposing kicker John Kasay. ...  Vikings would respond with another Longwell field goal (a 22-yard FG) ... Longwell booted the game-winning 19-yard field goal ''	& Question needed counting of certain events and none of these appeared as numbers. Model was able to apply count over number entities correctly \\ \hline

How many delegates were women from both the Bolshevik delegates and the Socialist Revolutionary delegates?	&  ``Of these mandatory candidates, only one Bolshevik and seven Socialist Revolutionary delegates were women.'' &	Model was able to apply sum on the correct numbers, even though many of the "how many" type questions need counting \\ \hline

How many years in a row did the GDP growth fall into negatives?	& ``Growth dropped to 0.3\% in 1997, -2.3\% in 1998, and -0.5\% in 1999.'' &	Model had to understand which numbers are "negative". It also needed to understand to count the two events instead of taking difference of the years \\ \hline

At it's lowest average surface temperature in February, how many degrees C warmer is it in May?	& ``The average surface water temperature is 26-28 C in February and 29 C in May.'' &	Passage had distrative unrelated numbers in the proximity but the model was able to select the lowest temperature out of (26,28) and then take difference of (29-26) \\ \hline

How many years ibefore the blockade was the Uqair conference taken place?	& ``Ibn Saud imposed a trade blockade against Kuwait for 14 years from 1923 until 1937... At the Uqair conference in 1922, ...	''	& Passage had other distracting unrelated numbers in the proximity but the model was able to select the correct difference operation \\ \hline
\end{tabular}
}
\caption{Manual Analysis of a few \emph{hard} instances (with Question and Relevant Passage Excerpt) where {\shortname} top-1 prediction was found to be correct}
\label{appendix:example_analysis}
\end{table} 

\newpage
\subsection{Background: Numerical Reasoning over Text}
\label{appendix:related_work}
The most generic form of Numerical reasoning over text (NRoT) is probably encompassed by the machine reading comprehension (MRC) framework (as in \cite{Dua2019DROP}), where given a long passage context, $c$, the model needs to answer a query $q$, which can involve generating a numerical or textual answer or selecting a numerical quantity or span of text from the passage or query. The distinguishing factor from general RC is the need to perform some numerical computation using the entities and numbers in the passage to reach the goal. 

\emph{Discrete/symbolic reasoning in NRoT}: In the early NRoT datasets \cite{hosseini-etal-2014-learning,roy-roth-2015-solving,koncel-kedziorski-etal-2016-mawps} which deal with simpler math word problems with a small context and few number entities, symbolic techniques to apply discrete operations were quite popular. However, as the space of operations grow or the question or the context becomes more open-ended these techniques fail to generalize. Incorporating explicit reasoning in neural models as discrete operations requires handling non-differentiable components in the network which leads to optimization challenges. 

\emph{Discrete reasoning using RL}: Recently Deep Reinforcement Learning (DRL) has been employed in various neural symbolic models to handle discrete reasoning, but mostly in simpler tasks like KBQA, Table-QA, or Text-to-SQL \cite{zhong2017seq2sql,NIPS2018_8204,liang2017neural,saha-etal-2019-complex,ijcai2019-679,DBLP:conf/iclr/NeelakantanLAMA17}. Such tasks can be handled by well-defined components or modules, with well structured function-prototypes (\ie function arguments can be of specific variable-types \eg KB entities or relations or Table row/column/cell values), which can be executed entirely as a symbolic process. On the other hand, MRC needs more generalized frameworks of modular networks involving fuzzy forms of reasoning, which can be achieved by \emph{learning} to execute the query over a sequence of learnable neural modules, as explored in \cite{Gupta2020Neural}. This was inspired by the Neural Modular Networks which have proved quite promising for tasks  requiring similar fuzzy reasoning like Visual QA \cite{DBLP:conf/cvpr/AndreasRDK16,DBLP:journals/corr/AndreasRDK15}.
\vspace{-1em}
\paragraph{SoTA models on DROP:} While the current leaderboard-topping models already showcase quite superior performance on the reasoning based RC task, it needs closer inspection to understand whether the problem has been indeed fully solved. 

\emph{Pre-trained Language Models}: On one hand, the large scale pretrained language models \cite{Geva2020InjectingNR} use Transformer encoder-decoder (with pretrained BERT) to emulate the input-output behavior, decoding digit-by-digit for numeric and token-by-token for span based answers. However such models perform poorly when only trained on DROP and need additional synthetic dataset of numeric expressions and DROP-like numeric textual problems, each augmented with the \emph{gold numeric expression} form.

\emph{Reasoning-free Hybrid Models}: On the other hand, a class of \emph{hybrid} neural models have also gained SoTA status on DROP by explicitly handling the different types of numerical computations in the standard extractive QA pipeline. Most of the models in this genre, like NumNet (\cite{ran-etal-2019-numnet}), NAQANet (\cite{Dua2019DROP}), NABERT+(\cite{nabert}), MTMSN (\cite{hu2019multi}) and NeRd (\cite{DBLP:conf/iclr/ChenLYZSL20}) do not actually treat it as a reasoning task; instead they precompute an exhaustive enumeration of all possible outcomes of numerical and logical operations (\eg \emph{sum/diff, negate, count, max/min}) and augment the training data with knowledge of the query-type (depending on reasoning-type) and \emph{all} the numerical expression that leads to the correct answer. This reduces the question-answering task to simply learning a multi-type answer predictor to classify into the reasoning-type and directly predict the numerical expression, thus alleviating the need for rationalizing the inference or handling any (non-differentiable) discrete operation in the optimization. 
Some of the initial models in this genre are NAQANet(\cite{Dua2019DROP} and NumNet (\cite{ran-etal-2019-numnet}) which are respectively numerically aware enhancements of QANet(\cite{wei2018fast}) and the Graph Neural Networks. These were followed by BERT-based models, NABERT and  NABERT+(\cite{nabert}), i.e. a BERT version of the former, enhanced with \emph{standard numbers} and \emph{expression templates} for constraining numerical expressions. MTMSN \cite{hu2019multi} models a specialized multi-type answer predictor designed to support specific answer types (e.g., count/negation/add/sub) with  supervision of the arithmetic expressions that lead to the gold answer, for each type. 

\emph{Modular Networks for Reasoning}: NMN \citep{Gupta2020Neural} is the first model to address the QA task through explicit reasoning by learning to execute the query as a specialized program over learnable modules tailored to handle different types of numerical and logical operations. However, to do so, it further needs to augment the training data with annotation of the \emph{gold program} and \emph{gold program execution} i.e. the \emph{exact} discrete operation and numerical expression (\ie the numerical operation and operands) that leads to the correct answer for \eg the supervision of the gold numerical expression in \Cref{fig:example} is \emph{\small{SUM(23, 26, 42)}}.  This is usually obtained through manual inspection of the data through regex based pattern matching and heuristics applied on the query language. However, because of the abundance of templatized queries in DROP this pattern matching is infact quite effective and noise-free, resulting in the annotations acting as strong supervision.  

However such a manual intensive process severely limits the overall model from scaling to more general settings. This is especially true for some of the previous reasoning based models, NABERT+, NumNet and MTMSN which perform better on than NMN (infact achieve SoTA performance) on the full DROP dataset. But we do not consider them as our primary baselines, as, unlike NMN, these models (\cite{hu2019multi,Efrat2019TagbasedME,Dua2019DROP,ran-etal-2019-numnet}) do not have any provision to learn in absence of the additional supervision generated through exhaustive enumeration and manual inspection.  
\citep{Gupta2020Neural} have been the first to train a modular network strong, \emph{albeit} a more fine-grained supervision for a fraction of training data, and auxiliary losses that allow them to learn from the QA pairs alone. Consequently on a carefully-chosen subset of DROP, NMN showcased better performance than NABERT and MTMSN, when strong supervision is available only for partial training data. 

Our work takes it further along the direction in two ways
\begin{itemize}
\vspace{-0.5em}
    \item while NMN baseline can handle only 6 specific kinds of reasoning, for which they tailored the program generation and gold reasoning annotation, our model works on the full DROP-\emph{num}, that involves more diverse kinds of reasoning or more open-ended questions, and requires evaluating on a subset $\times$7.5, larger by training on $\times$4.5 larger training data.
    \vspace{-0.2em}
    \item while NMN generalized poorly on the full DROP-\emph{num}, especially when only one or more types of supervision is removed, our model performs significantly better without any of these types of supervision.
\end{itemize}
\vspace{-0.5em}
Together, NMN and GenBERT are some of the latest works in the two popular directions (reasoning and language model based) for DROP that allow learning with partial no strong supervision and hence act as primary baselines for our model.

Since in this work we are investigating how neural models can incorporate explicit reasoning, we focus on only answering questions having numerical answer (DROP-\emph{num}), where we believe the effect of explicit reasoning is more directly observeable. This is backed up by the category-wise performance comparison of reasoning-free language model GenBERT (reported in \cite{Geva2020InjectingNR}) with other hybrid models (MTMSN and NABERT+) that exploit numerical computation required in answering DROP questions. While, on DROP-\emph{num}, there is an accuracy gap of 33\% between the GenBERT model and the hybrid models (when all are trained on DROP only), there is only a 2-3\% performance gap on the subset having answers as single span, despite the latter also needing reasoning. This evinces that the performance gap is indeed due to exploiting explicit reasoning under such strong supervised settings.

\subsubsection{Limitations of NMN}
\label{appendix:nmn_limitations}
The primary motivation behind our work comes from some of the limitations of the contemporary neural module networks, NMN and the reasoning-free hybrid models MTMSN, NABERT+, NumNet, NAQANet; specifically their dependence on the availability of various kinds of strong supervision. For that we first describe the nature of programmatic decompositions of queries used in the modular architectures in the closest comparable work of NMN.

NMN defined a program structure with modules like `find', `filter', `relocate', `find-num', `find-date', `year-difference', `max-num', `min-num', `compose-number' etc., to handle a carefully chosen subset of DROP showcasing only 6 types of reasoning,  (i.e. \emph{Date-Difference, Count, Extract Number, Number Compare}). For e.g. for the query \emph{Which is the longest goal by Carpenter?} the program structure would be {\small\emph{(MAX(FILTER(FIND(`Carpenter'), `goal'))}}, where each of these operations are learnable networks. However to facilitate learning of such specialized programs and the networks corresponding to these modules, the model needs precomputation of the exhaustive output space for different discrete operation and also various kinds of strong supervision signals pertaining to the program generation and execution. 

 \emph{Precomputation of the Exhaustive Output-Space}: For operations that generate a new number as its output (\eg \emph{sum/diff}), the annotation enumerates the set of all possible outputs by computing over all subsets of number or date entities in the passage. This simplifies the task by allowing the model to directly learn to optimize the likelihood of the arithmetic expression that lead to the final answer, without any need for handling discrete operations. 
 
\emph{Program Supervision} provides supervision of the query category out of the 6 reasoning categories, on which their program induction grammar is tailored to. With this knowledge they can directly use the category specific grammar to induce the program ( for e.g.  \emph{\small{SUM(FILTER(FIND))}} in Fig \ref{fig:example}). Further all these models (NMN, MTMSN, NABERT+, NumNet, NAQANet)  use the supervision of the query category to understand whether the discrete operation is of type count or add/sub or max/min.  which includes the knowledge of the `gold' discrete operation (i.e. count or max/min or add/sub) to perform.

\emph{Query Attention Supervision} provides information about the query segment to attend upon in each step of the program, as the program argument for e.g. in Fig \ref{fig:example}, \emph{\small{`Carpenter'}} and \emph{\small{`goal'}} in the 1st and 2nd step of the program.

\emph{Execution Supervision}: For operations that select one or more of the number/date entities in the passage, (for e.g. max/min), rule based techniques provide supervision of the subset of numbers or dates entities from the passage, over which the operation is to be performed.

These annotations are heuristically generated through manual inspection and regular expression based pattern matching of queries, thus limiting their applicability to a small subset of DROP only. Furthermore, using a  hand-crafted grammar to cater to the program generation for each of their reasoning categories, hinders their generalizability to more open ended settings. While this kind of annotation is feasible to get in DROP, this is clearly not the case with other futuristic datasets, with more open-ended forms of query, thus calling for the need for other paradigms of learning that do not require such manually intensive annotation effort.

\subsubsection{Pretraining Data for GenBERT} 
\label{appendix:genbert_pretraining}
\begin{wrapfigure}{r}{0.3\textwidth}
\vspace{-1em}
\includegraphics[width=\linewidth]{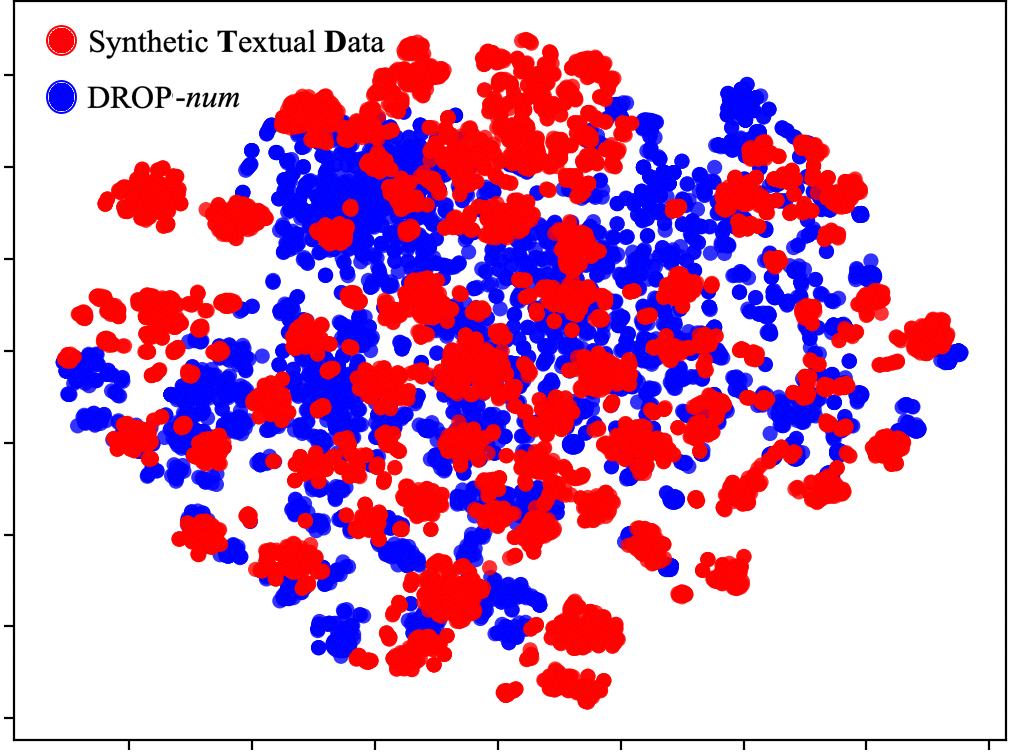}
\vspace{-1em}
\caption*{Figure 6: t-SNE of questions in DROP-\emph{num}-Test and Synthetic Textual Data used in GenBERT models (TD and ND+TD)}
\label{fig:plot_td}
\end{wrapfigure}
While GenBERT (\cite{Geva2020InjectingNR}) greatly benefits  from pretraining on synthetic data, there are few notable aspects of how the synthetic textual data was carefully designed to be similar to DROP. The textual data was generated for the same two categories \emph{nfl} and \emph{history} as DROP with similar vocabulary and involving the same numerical operations over similar ranges of numbers (2-3 digit numbers for DROP and 2-4 digit numbers for synthetic textual data). The intentional overlap between these two datasets is evident from the t-SNE plots (in Figure 6) of the pretrained Sentence-Transformer embedding of questions from DROP-\emph{num} (blue) and the Synthetic Textual Data (red). Further, while the generalizability of GenBERT was tested on add/sub operations from math word problems (MWP) datasets ADD-SUB, SOP, SEQ, their synthetic textual data was also generated using the same structure involving world state and entities and verb categories used by \cite{hosseini-etal-2014-learning} to generate these MWP datasets. Such bias limits mitigates the real challenges of generalizability, limiting the true test of  robustness of such language models for numerical reasoning. 

\begin{figure*}[!htb]
\centering
\includegraphics[width=\linewidth]{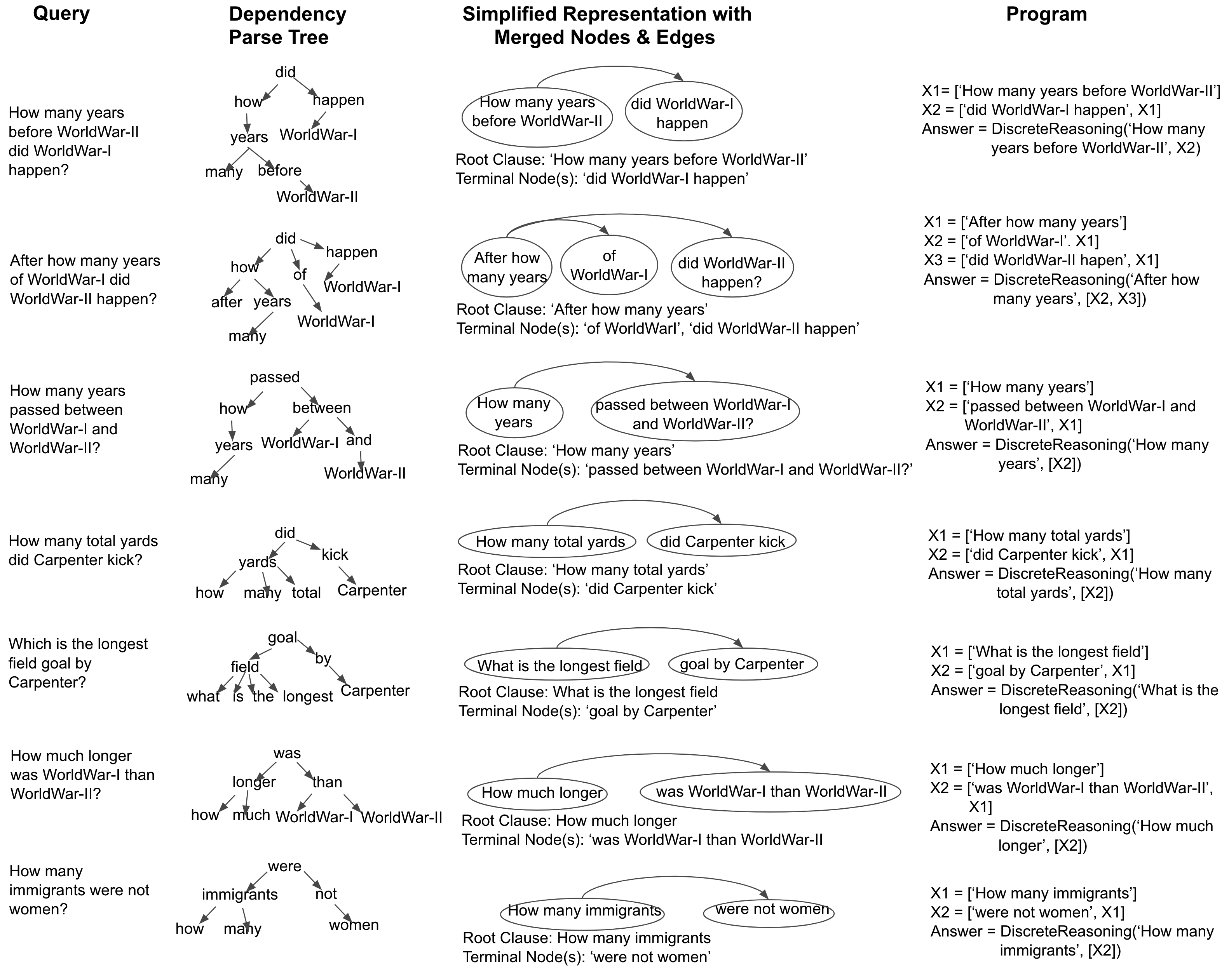}
\caption*{Figure 7: Examples of Programs for {\shortname} obtained from the Dependency Parse Tree of the Query}
\label{fig:parses}
\end{figure*} 
\subsection{Query Parsing: Details} 
\label{appendix:query_parse}
The Stanford Dependency parse tree of the query is organized into a program structure as follows
\begin{itemize}
    \item \textbf{Step 1)} A node is constructed out of the subtrees rooted at each immediate child of the root, the left-most node is called the root-clause
    \item \textbf{Step 2)} Traversing the nodes from left to right, an edge is added between the left-most to every other node, and each of these are added as steps of the program with the node as the query span argument of that step and the reference argument as the incoming edges from past program steps
    \item \textbf{Step 3)} The terminal (leaf) nodes obtained in this manner are then further used to add a final step of the program which is responsible for handling the discrete operation. The query-span argument of this step is the root-clause, which often is indicative of the kind of discrete reasoning to perform. The reference arguments of this step are the leaf nodes obtained from Step 2).
\end{itemize}
Figure 7 provides some example queries similar to those in DROP along with their Dependency Parse Tree and the Simplified Representation obtained by constructing the nodes and edges as in Step 1) and 2) above, and the final program which is used by {\shortname}. 
Note that in this simplified representation of the parse tree the root-word of the original parse tree is absorbed in its immediate succeeding child. Also we simplify the structure in order to limit the number of reference arguments in any step of the program to 2, which in turn requires the number of terminal nodes (after step 2 of the above process) to be limited to 2. This is done in our left to right traversal by collapsing any additional terminal node into a single node.

\subsection{RL Framework: Details}
\label{appendix:RL}
In this section we discuss some additional details of the RL framework and tricks applied in the objective function\\
\textbf{Iterative ML Objective}: In absence of supervision of the true discrete action that leads to the correct answer, this iterative procedure fixes the policy parameters to search for the \emph{good} actions (where $\mathcal{A}^{good} = \{a: R(x, a)=1\}$) and then optimizes the likelihood of the \emph{best} one out of them. However, the simple, conservative approach of defining the best action as the most likely one according to the current policy can lead to local minima and overfitting issues, especially in our particularly sparse and confounding reward setting. So we take a convex combination of a conservative and a non-conservative selection that respectively pick the most and least likely action according to the current policy out of $\mathcal{A}^{good}$ as best. Hyperparameter $\lambda$ weighs these two parts of the objective and is chosen to be quite low $(1e^{-3})$, to serve the purpose of an epsilon-greedy exploration strategy without diverging significantly from the current policy.
\vspace{-1em}
\begin{dmath*}
    J^{IML}(\theta,\phi) = \sum_x (1-\lambda)  \max_{a\in \mathcal{A}^{good}} \log{P_{\theta,\phi} (a|x)} + \lambda \min_{a\in \mathcal{A}^{good}} \log{P_{\theta,\phi} (a|x)} 
\end{dmath*}
\vspace{-1em}
\textbf{Using Noisy Pseudo-Reward}: In addition to using the REINFORCE objective to maximise the likelihood of actions that lead to the correct answer, we can also obtain different noisy \emph{pseudo rewards} ($\in \{-1, +1\}$) for the different modules that contribute towards the action sampling (i.e. the operator and the entity-type and different argument sampler networks). Towards this end, we define pseudo-reward for sampling an operator as the maximum of the reward obtained from \emph{all} the actions involving that operator. Similarly, we can also define reward for predicting the entity-type (date or number) over which the discrete operation should be executed. Following the same idea, we also obtain pseudo rewards for the different argument sampling modules. For e.g. if the most likely operator (as selected by the Operator Sampler) is of type \texttt{count} and it gets a pseudo-reward of $+1$, then, in that case, we can use the reward obtained by the different possible outputs of the Counter network as a noisy pseudo-label supervision and subsequently add an explicit loss of negative log-likelihood to the final objective for the Counter module. Similar pseudo-reward can be designed for the Entity-Ranker module when the most likely operator sampled by the Operator Sampler needs arbitrary number of arguments.  Treating the pseudo-reward as a noisy label can lead to a negative-log-likelihood based loss on output distribution from the Entity-Ranker, following the idea that the correct entities should atleast be ranked high so as to get selected when sampling any arbitrary number of entities.

\end{document}